%% file: main.tex
\RequirePackage{tikz}
\documentclass[default,iicol,sn-mathphys]{sn-jnl}
\jyear{2022}

\usepackage[ruled,vlined,linesnumbered,algo2e]{algorithm2e}
\usepackage{aircraftshapes}
\usepackage{caption}

\usepackage{multirow}
\usepackage{import}
\subimport{./}{init}

\usetikzlibrary{positioning}
 \usetikzlibrary{3d}

\newcommand{\cR}{{\cal Q}}

\theoremstyle{thmstyleone}%

\newtheorem{definition}{Definition}

\begin{document}

\title[Energy-constrained Coverage Learning]{Deep Recurrent Q-learning for Energy-constrained\\ Coverage with a Mobile Robot}
\author[1]{\fnm{Aaron} \sur{Zellner}}\email{n01185806@unf.edu}

\author*[1]{\fnm{Ayan} \sur{Dutta}}\email{a.dutta@unf.edu}

\author[1]{\fnm{Iliya} \sur{Kulbaka}}\email{n01427009@unf.edu}

\author[2]{\fnm{Gokarna} \sur{Sharma}}\email{gsharma2@kent.edu}

\affil[1]{\orgdiv{School of Computing}, \orgname{University of North Florida}, \orgaddress{ \city{Jacksonville}, \postcode{32224}, \state{FL}, \country{USA}}}

\affil[2]{\orgdiv{Department of Computer Science}, \orgname{Kent State University}, \orgaddress{ \city{Kent}, \postcode{44242}, \state{OH}, \country{USA}}}




\abstract{
In this paper, we study the problem of coverage of an environment with an energy-constrained robot in the presence of multiple charging stations. As the robot's on-board power supply is limited, it might not have enough energy to cover all the points in the environment with a single charge. Instead, it will need to stop at one or more charging stations to recharge its battery intermittently. The robot cannot violate the energy constraint, i.e., visit a location with negative available energy. To solve this problem, we propose a deep Q-learning framework that produces a policy to maximize the coverage and minimize the budget violations. Our proposed framework also leverages the memory of a recurrent neural network (RNN) to better suit this multi-objective optimization problem. We have tested the presented framework within a $16 \times 16$ grid environment having charging stations and various obstacle configurations. Results show that our proposed method finds feasible solutions and outperforms a comparable existing technique.
}


\maketitle

\section{Introduction}
Path planning has been in the center of attention in the mobile robotics research domain for the past several decades. In a shortest path planning (SPP) problem, given a map of the environment, the goal of the robot is to plan a path from the start to the goal location that minimizes a pre-defined cost function~\cite{dijkstra1959note,hart1968formal,madkour2017survey}. This has applications in route planning in automated taxi services, among others. On the other hand, in a coverage path planning (CPP) setting, the goal of the robot is to go through all the locations in the environment at least once. In this case, although there is a start point, there is no specific goal location given~\cite{cabreira2019survey,choset2001coverage,galceran2013survey}. Applications of CPP include automated wall painting and vacuum cleaning, among others. In this paper, we study CPP and not SPP.

Only recently, researchers have started looking at a more practical variant of these classic path planning problems, i.e., where the robot  is cordless meaning that it is battery operated and  has a limited battery budget~\cite{sharma2019optimal,shnaps2016online,sotolongo2021shortest,sundar2013algorithms}. Most modern day ground mobile robots cannot work for more than an hour or two before requiring to recharge its battery. Therefore, it is imperative that we consider the energy constraint put on by the robot's on-board battery. Let us take a running example of CPP under the energy constraint. Let us assume that an iRobot Roomba is vacuum cleaning a house and there are multiple charging stations located in it. To clean the house completely, the objective of the Roomba robot would be to go to one of the charging stations before running out of battery while making sure that all the points on the floor are cleaned. As the user would have to manually put the robot on to a charging station if it runs out of battery anywhere else, we would have to minimize the number of such energy constraint violations occur. CPP is related to a neighborhood variant of the famous Travelling Salesman Problem (TSP), where the objective is to visit the neighborhoods of the cities~\cite{arkin1994approximation}. Due to the NP-hardness of TSP, the complexity of this problem also increases drastically with higher dimensions. A specific type of CPP, called the lawn-mowing pattern planning for grass cutting, has been proved to be NP-hard even in an obstacle-free environment~\cite{arkin2000approximation}. To avoid such complexities, vacuum cleaning robots, such as Roombas, randomly move in the environment for long periods conjecturing that all the dirt will be collected~\cite{palacin2005measuring}. 

\begin{figure}[ht!]
    \centering
    \includegraphics[width=0.7\linewidth]{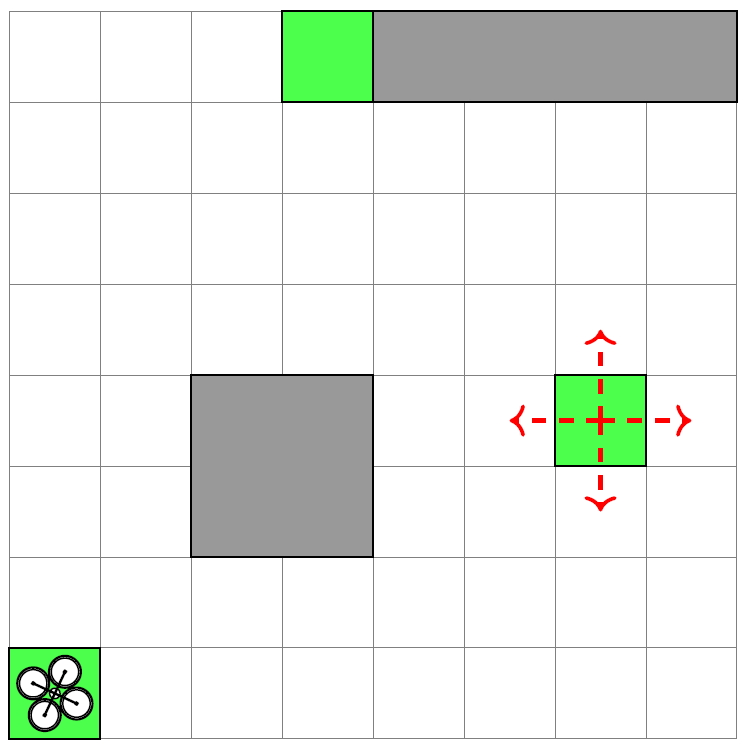}
    \caption{An illustration of a typical rectangular environment. The green color indicates the charging stations and gray indicates obstacles. The robot always starts from a charging station as shown here. Four maximum possible actions from any cell are shown with red dashed lines with arrows.}
    \label{fig:illust}
\end{figure}

To this end, we propose a deep Q-learning framework, which neither requires careful algorithm planning to address the complexity bottleneck, nor does it require any prior domain knowledge. Instead, it learns over time to maximize the coverage, i.e., the number of unique locations visited by the robot, and to minimize the number of energy constraint violations -- the number of non-charging locations the robot reaches with no battery energy remaining. The spatial information is extracted using two convolutional neural layers. As the robot's past coverage helps the robot plan a more informed path in the future, it is imperative to take past decisions into account. To do this, we use a Long Short Term Memory (LSTM) network -- a type of recurrent neural network (RNN), which allows information from the past experiences to persist. Similar to a standard vacuum cleaning robot, we assume our robot to have an on-board obstacle detection sensor (e.g., infrared) to detect and avoid obstacles. This comes in handy if the obstacle locations in the environment are not known \textit{a priori}. We have implemented the proposed framework in Python and tested it within a $16 \times 16$ four-connected grid environment with different obstacle configurations. We have also tested with varied battery budget amounts. Results show that our proposed technique learns paths for the robot to cover the entire environment without violating the budget constraint. 

In summary, the primary contributions of our paper are as follows.
\begin{itemize}
    \item To the best of our knowledge, this is the first study that formulates the energy-constrained CPP problem as a deep Q-learning problem.
    \item We have used a recurrent layer along with convolutional layers for the robot to learn from the history of spatial navigation. 
    \item Our proposed learning model has been empirically shown to learn effective coverage paths for the robot to completely explore environments with various obstacle configurations and battery budget amounts. 
\end{itemize}

\section{Related Work}
CPP is one of the most well-known research problems in mobile robotics. This has applications in painting~\cite{sheng2000automated,zhou2022building}, agriculture~\cite{dutta2021multi,pham2017aerial,valente2013aerial}, mapping~\cite{almadhoun2019survey,matignon2018multi,paull2018probabilistic}, among others. Surveys of CPP techniques in robotics can be found in ~\cite{cabreira2019survey,choset2001coverage,galceran2013survey}. CPP is similar to the covering salesman problem (CSP)~\cite{current1989covering}, which is a variant of the well-known NP-hard travelling salesman problem (TSP). Covering an environment even without any obstacle has been proved to be NP-hard~\cite{arkin2000approximation}. In this paper, we are interested in grid-based coverage, where the environment is divided into uniform cells and binary values embedded into these cells indicate the presence or absence of obstacles. Wavefront planners can be used for covering such discrete environments~\cite{zelinsky1993planning}. Other techniques involve spanning tree~\cite{gabriely2002spiral} and neural network~\cite{yang2004neural}-based solutions. However, most of these exiting solutions in the literature assume the robot to have unlimited on-board power supply. Only recently, the researchers have started considering the energy constraint of the robot~\cite{dogru2022eco,sharma2019optimal,shnaps2016online,WeiI18}. In this setting, due to the limited energy budget, the robot will need to stop at one or more charging stations to cover the entire environment. Understandably, this adds a new challenge to the already complex CPP with a mobile robot. In \cite{shnaps2016online}, the authors have shown that in an online setting, where the obstacle patterns and locations are unknown to the robot, the optimal coverage can only provide $O(\log B)$ approximation. Sharma et al. \cite{sharma2019optimal} have presented the first such online algorithm. For the offline version, where the environment is known (similar to our model in this paper), Wei and Isler presented 
$O(\log B)$-approximation~\cite{WeiI18} and constant-factor approximation~\cite{WeiICRA18} algorithms.

On the other hand, due to the advancement in the field of deep learning and reinforcement learning, some recent studies have employed deep reinforcement learning (DRL) techniques for CPP with a mobile robot. The benefit of such an approach is that no domain knowledge and/or careful designing of combinatorial algorithm is needed. Examples include ~\cite{apuroop2021reinforcement,fevgas2022coverage,kyaw2020coverage,lakshmanan2020complete,zhu2019complete}. Among these, the study by Zhu et al.~\cite{zhu2019complete} consider CPP with an underwater robot. Similar to our paper, Lakshmanan et al.~\cite{lakshmanan2020complete} and Apuroop et al.~\cite{apuroop2021reinforcement} have used convolutional layers along with a recurrent module for complete area coverage. Fevgas et al.~\cite{fevgas2022coverage} have proposed an energy-efficient coverage technique for multi-robot CPP. Kyaw et al.~\cite{kyaw2020coverage} have presented an attention-based actor-critic neural network architecture for CPP. Unlike our studied problem in this paper, none of these DRL-based solutions from the literature consider the energy constraint of the robot and the presence of charging stations in the environment.







\section{Model}
\noindent{\bf Environment. } The environment $P$ is a planar polygon containing a set of charging stations $S = s_1, s_2, \cdots, s_k$ inside it. 
$P$ may possibly contain polygonal obstacles. The obstacles are assumed to be static meaning that they do not move during the coverage process. See Fig.~\ref{fig:illust} for an illustration of $P$ with two polygonal obstacles. $P$ is discretized into cells forming a 4-connected grid. 

\smallskip
\noindent{\bf Robot. } We consider the robot $r$ to be initially positioned at a charging station $s_s$ inside $P$. $r$ has size $L\times L$ that fits within a grid-cell in $P$. 
The robot $r$ can move to any of the four orthogonal neighbor cells (if the cell is not occupied by an obstacle) from its current cell. We also assume that $r$ has the knowledge of the global coordinate system through a compass on-board, that means it knows left (West), right (East), up (North), and down (South) cells consistently from its current cell. 
$r$ is equipped with a position sensor (e.g., GPS) and an obstacle-detection sensor (e.g., laser rangefinder). We assume that with the laser rangefinder, the robot can detect obstacles in any of its neighbor cells. Let $p_{crnt}$ and $p_{nxt}$ denote $r$'s current and next locations in $P$. $neigh(p)$ represents the maximum of $4$ neighbor cells of any cell $p \in P$ and $p_{nxt} \in neigh(p_{crnt})$. The robot has sufficient on-board memory to store information necessary to facilitate the coverage process. The energy consumption of the robot is proportional to the distance travelled, i.e., the energy budget of $B$ allows the robot to move $B$ units distance.

\smallskip\noindent
\textbf{Objective Function. } 
A path $\cR$ is a list of grid-cells that $r$ visits starting with $s_s$. Let $br(p)$ denote the remaining battery budget at cell $p \in P$. Notice that if there are some obstacles within $P$ located in such a way that they divide $P$ into two sub-polygons $P_1$ and $P_2$ with $P_1$ and $P_2$ sharing no common boundary, then $r$ cannot fully cover $P$. Therefore, it is considered that there is no such cell $p \in P$, i.e., there is (at least) a path from some $s_i \in S$ to any obstacle-free cell of $P$. We call a cell {\em free} if it is not occupied by an obstacle. 

\begin{definition}[Reachable Cell]
\label{definition:reachable}
Any cell $p \in P$ is called {\em  reachable} by the robot $r$, if and only if 
(a) it is a free cell,
(b) it is within distance $\lfloor B/2\rfloor$ from a charging station $s_i \in S$, and 
(c) there must be at least a path of consecutive free cells from $s_i$ to $p$.
\end{definition}

The goal of the coverage algorithm is to find a path $\cR$ for the robot such that 
\begin{itemize}
\item {\bf Condition (a):} $br(p_q) \geq 0$ for each cell $p_q \in \cR$. 
\item {\bf Condition (b):} The cells in $\cR$ collectively cover the whole environment $P$, i.e., $\cup_{\vert \cR \vert} \cR=P$, 
\end{itemize}

and the following performance metric is optimized:  
\begin{itemize}
\item {\bf Performance metric:} The {\em total length of the path} in $\cR$, denoted as $l(\cR)$, is minimized. 
\end{itemize}

Note that we do not consider the charging time of the battery since this delay does not impact the performance metrics we consider. The coverage path planning problem is formally defined as follows.
\begin{definition}
Given a planar polygonal environment $P$ possibly containing obstacles, and a mobile robot $r$ having battery budget of $B$ initially positioned at a charging station $s_s$ inside $P$, the objective of $r$ is to visit all the reachable cells in $P$ through a path  such that 
\begin{itemize}
\item Conditions (a) and (b) are satisfied, and 
\item The performance metric is minimized.
\end{itemize}
\end{definition}

\section{Recurrent Q-learning for Energy-constrained CPP}

\begin{figure}
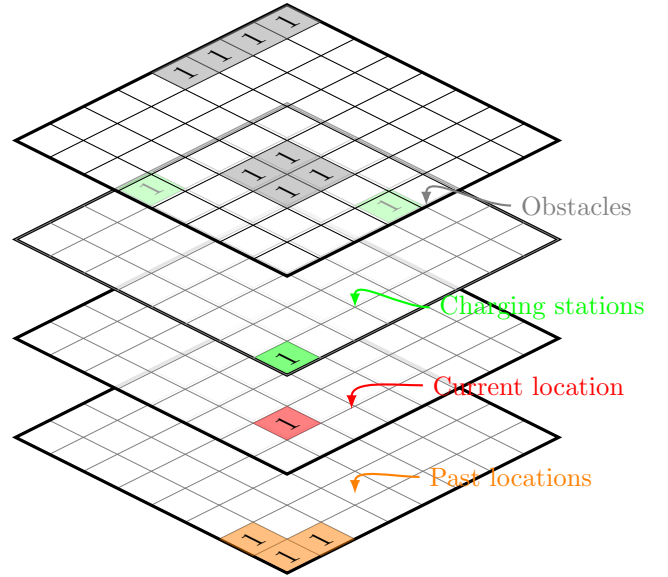

    \centering
    \include{statespace_illust}
    \caption{Inputs to our proposed neural network model are shown (following the illustrated environment in Fig. \ref{fig:illust}) -- the obstacle and charging station matrices do not change over time. However, the current location matrix is updated after every action is taken by the robot and a new cell is reached. The empty cells in the matrices are filled with 0's (not shown here for brevity).}
    \label{fig:statespace}
\end{figure}

\subsection{State, Action, and Reward} 
\noindent
\textbf{State Space. } A state $x \in \mathcal{S}$ of the robot is represented by a $4 \times N \times N$ tensor (simply a 4D tensor), where $N$ is the length of one side of the environment.

\textit{Obstacle channel. } The obstacles in the environment are represented with $1$'s, i.e., the cells that are covered with obstacles contain $1$'s and the free cells contain $0$'s.

\textit{Charging station channel. } The locations of the charging stations $S$ are represented with $1$'s and the grid cells that are not charging stations contain $0$'s. 

\textit{Current location channel. } The next layer represents the robot's current grid cell location -- it is denoted by a $1$ whereas the remaining cells contain $0$'s. 

\textit{Covered locations channel. } The state also represent the grid cell locations that the robot has visited in the past. These visited cells are denoted by $1$'s and the non-visited cells are denoted by $0$'s. Note that even if a cell is visited more than once, it is still represented by a $1$. 

An example of the state space is shown in Fig. \ref{fig:statespace}. The idea of using multi-dimensional binary tensors as input states is popular in the literature \cite{heydari2021reinforcement,theile2020uav}.

\smallskip
\noindent
\textbf{Action Space. } The action space, $\mathcal{A}$, of the robot contains four perpendicular actions: \{Up, Down, Left, Right\}.

\smallskip
\noindent
\textbf{Reward Function. } The reinforcement learning is driven by the underlying reward structure. A reward function $R: \mathcal{S} \times \mathcal{A} \longrightarrow \mathbb{R}$ maps a state-action pair $(s,a)$ to a real number. In this paper, we model the reward such that our multi-objective optimization criteria is accounted for. Let $r^p$ and $r^s$ denote the rewards for our two optimization criteria respectively: 1) maximize the number of visited cells, and 2) minimize the number of energy constraint violations. These individual reward functions are defined as follows where $p_i \in P$ is the cell where $r$ reaches after executing $a\in \mathcal{A}$ in state $x \in \mathcal{S}$.

\begin{equation}
   r^p(p_i) = 
    \begin{cases}
        2, & \text{if }
       \begin{aligned}[t]
            \text{$p_i$ is not visited before}
       \end{aligned}
        \\
        -1, & \text{otherwise}\\
\end{cases}
\end{equation}

\begin{equation}
   r^s(p_i) = 
    \begin{cases}
        0.1, & \text{if }
       \begin{aligned}[t]
            br(p_i) \geq 0
       \end{aligned}
        \\
        -3, & \text{otherwise}\\
\end{cases}
\end{equation}

To combine these two sub-reward components, we employ a weighted linear combination technique as follows.

\begin{equation}
    R = \frac{r^p (p_i) + r^s (p_i)}{2}
    \label{eq:reward}
\end{equation}

Following Eq. \ref{eq:reward}, the robot will try to visit the unvisited grid cells without violating the budget than to repeatedly visit the already visited cells. A high negative reward of $-3$ will also lead to a policy where the number of energy constraint violations reduce. If the robot completes covering the whole environment without any violation, for the last action it takes, we give it a termination reward of $R = 200$. 

\subsection{Background on Reinforcement Learning}
In reinforcement learning, the robot interacts with an environment in discrete time steps $T$. First it takes an action $a \in \mathcal{A}$ in state $x$, receives a corresponding reward $R$, and the state of the robot changes to $x' \in \mathcal{S}$ because of the action execution. The process continues until a terminal state is reached. Next, the environment is reset and the above-mentioned steps are repeated. The expected value of taking action $a$ in state $x$ is given by its Q-value: $Q(x,a) = \sum\limits_{k=0}^{\infty} \gamma R_{t+k}$. $\gamma \in [0,1]$ is the discount factor that balances the importance of present and future rewards in the Q-value. An optimal policy can be found in this one-step setting by taking the highest Q-value actions in each state: $Q^*(x,a) = \max Q(x,a)$. Model-free Q-learning is an example of one type of reinforcement learning to learn the optimal action Q-values in each state~\cite{sutton1998introduction,watkins1989learning}. As it is infeasible to learn the optimal Q-values for all state-action pairs, the Q-function is recursively updated as the following.

\begin{multline}
    \label{eq:q_update}
    Q(x,a) = (1-\alpha)Q(x,a) + \\
    \alpha(R_t + \gamma \max_{a'\in \mathcal{A}} Q(x',a'))
\end{multline}

In the recent years with the advancements in deep learning, the Q-values are approximated using a neural network~\cite{mnih2015human,van2016deep}. The parameters in this network are denoted by $\phi$ and the network is represented by $Q_{\phi}$. The goal of this network is to output the Q-values of the state-action pairs. We adjust $\phi$ for $Q_{\phi}$ to output optimal Q-values.

\begin{figure*}
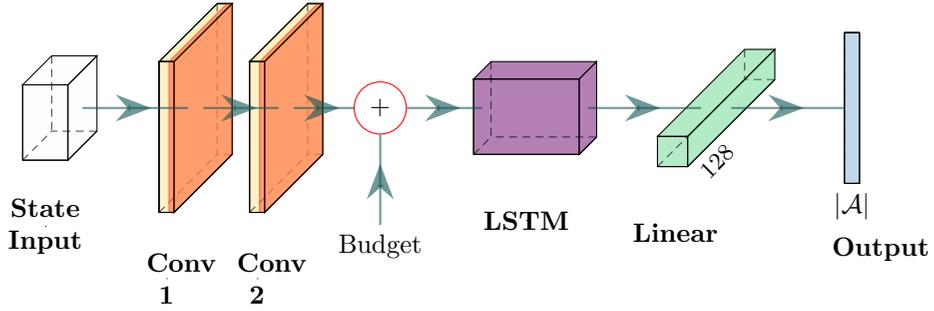

    \centering
    \include{NN_model}
    \caption{Deep recurrent Q-learning architecture for energy-constrained CPP.}
    \label{fig:model}
\end{figure*}

\subsection{Neural Network Architecture and Algorithm}
The proposed Q-network architecture is shown in Fig. \ref{fig:model}. The input to the network is a four channel state information, $x \in \mathcal{S}$. 
Note that in case of a no-obstacle environment, the last channel input will not be required. These inputs are then passed through two convolutional layers along with batch-normalization and rectified linear unit (ReLu) activation. We have used convolutional layers to extract the spatial information from the input states. Both the convolutional layers have $16$ output channels. The filter size and stride are set to $5$ and $1$ respectively in our experiments. The output of the second convolutional layer is concatenated with the remaining budget of the robot. Next, this is passed to an recurrent neural network (RNN) module~\cite{hausknecht2015deep}. We have used Long Short Term Memory (LSTM) as our RNN model~\cite{hochreiter1997long}. LSTM networks are useful for remembering long history of information such as time-series data. LSTM maintains a cell state that regularizes data addition or deletion using a gate. The hidden cell size of the LSTM module is set to $128$. The output of the LSTM is then given as an input to a linear layer of neurons, which has an output size of $\vert\mathcal{A}\vert$. These outputs are the Q-values of the $\vert\mathcal{A}\vert$ actions that the robot can take. If any action is infeasible, e.g., if it leads to a collision with an obstacle, we mask the corresponding Q-value with $-\infty$.



\begin{algorithm2e}
Initialize a priority replay memory $\mathcal{D}$, policy and target networks $Q_{\phi} \text{ and } Q_{\phi^-}$ respectively.\\
\For{each training episode $e$}{
    \For{each step $t$ in $e$}{
            Sample action $a \in \mathcal{A}$ using $\epsilon$-greedy strategy;\\
            Execute action $a$, receive reward $R$, and transition to state $x'$;\\
            Store transition information $\langle x,a, R,x' \rangle$ in replay buffer $\mathcal{D}$;
            }
            Sample a minibatch of $m$ experiences from $\mathcal{D}$;\\
            Calculate the target value $Y_Q$ (Eq. \ref{eq:target});\\
            Regress the $Q$-network towards this target value following Eq. \ref{eq:loss};\\
            Update the target network parameters $\phi^{-}$ every $\tau$ episodes;
        }
\caption{Deep Recurrent Q-learning for Energy-constrained CPP}
\label{algo}
\end{algorithm2e}

We present the pseudo-code of the proposed method in Algorithm \ref{algo}. Similar to Double DQN (DDQN)~\cite{van2016deep}, we maintain two copies of the neural network: one for the policy ($Q_{\phi}$) and the other for the target ($Q_{\phi^-}$) for better stabilization and reducing overestimation in action values~\cite{mnih2015human,van2016deep}. The training happens in episodes. In each episode, the robot starts from a starting cell $S$ and moves for $T$ steps. If the environment is fully explored before $T$ steps, the episode is terminated and the environment is reset. In each such step within an episode, $r$ first samples an action $a \in \mathcal{A}$ using the $\epsilon$-greedy strategy, i.e., it chooses the action with the maximum Q-value with probability $\epsilon$, otherwise $a$ is random with probability $(1-\epsilon)$. Next, it executes $a$ and receives a corresponding reward $R$ following Eq. \ref{eq:reward}. Due to the action execution, the current state of the robot changes from $x$ to $x'$. The experience tuple $\langle x,a, R,x' \rangle$ is stored into a memory replay buffer $\mathcal{D}$. We have employed a priority memory buffer~\cite{fedus2020revisiting}. In every step, a mini-batch of experience memories $\mathcal{B}$ is selected from $\mathcal{D}$ and the network parameters are regressed toward their target values by finding the gradient descent of the following temporal loss function

\begin{equation}
\label{eq:loss}
    \mathcal{L}({\phi}) = \mathbb{E}[(Y_Q(x,a,x')-
    Q_{\phi}(x,a))^2]
\end{equation}
\begin{equation}
\label{eq:target}
    Y_Q(x,a,x') = R + \gamma Q_{\phi^-}(x', \text{arg}\max Q_{\phi}(x',a'))
\end{equation}

Following DDQN, the best action $a$ is chosen using $Q_{\phi}$ but the expected Q-value of that action is estimated using $Q_{\phi^-}$~\cite{van2016deep}. The optimization is performed using the Adam optimizer after every action. On the other hand, the parameters of the target network (${\phi}^-$) are updated after every $\tau$ episodes. This continues until the pre-defined maximum number of episodes is reached. As the environment is assumed to be known, we keep track of the best coverage path found by the robot in training and use it as the final solution.

\section{Experiments}
\subsection{Settings}
\begin{figure*}[ht!]
    \centering
    \begin{tabular}{cccc}
    \includegraphics[width=0.3\linewidth]{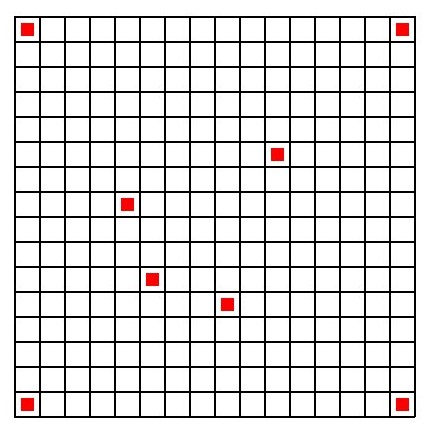} &
    \includegraphics[width=0.3\linewidth]{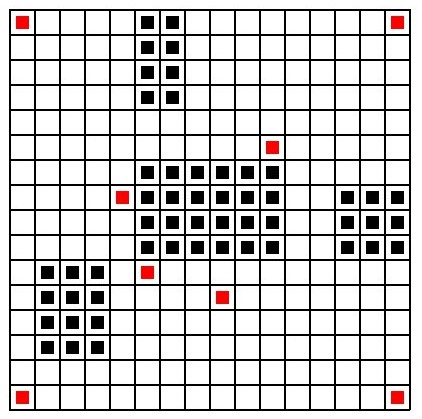}&
    \includegraphics[width=0.3\linewidth]{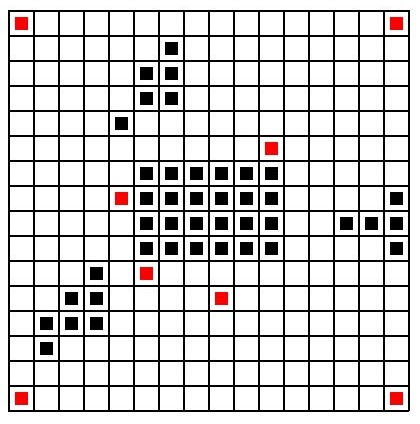}\\
    Map 1 & Map 2 & Map 3\\ 
    \end{tabular}
    \caption{Three different maps used in our experiments. Red cells indicate the charging stations and the black cells indicate the obstacles.}
    \label{fig:maps}
\end{figure*}

We have implemented the proposed algorithm in Python and used PyTorch library for deep reinforcement learning functions. The experiments are run on desktop computers with NVIDIA GPU and CUDA. The main parameters used in the experiments are listed in Table \ref{tab:params}. We have designed three environment configurations with charging stations and obstacles. These are named `Maps' and are numbered from 1 through 3 as shown in Fig. \ref{fig:maps}. The environment size is $16 \times 16$, i.e., $N=16$. The robot always starts from the $(0,0)$ corner and the budget of it is set to $5N$ unless mentioned otherwise. We have used a prioritized memory replay to store the experience transitions~\cite{SchaulQAS15}.

\textit{Baseline.} As mentioned earlier, this is the first work that studies mobile robot coverage under energy constraint in the presence of multiple charging stations. Therefore we could not compare against any existing baseline. Instead, we have used an existing CNN-based architecture as a baseline neural network model~\cite{theile2020uav}. This said CNN model has been developed for CPP with a unmanned aerial vehicle (UAV), where the UAV learns to cover an area with a limited energy budget. However, no charging station is available in the environment. Similar to our model, the map of the environment is passed to the CNN through multiple channels in this existing model~\cite{theile2020uav}. We call this baseline `CNN' in the remainder of the paper.

\begin{table}[ht!]
    \centering
    \begin{tabular}{ |p{4 cm} p{3 cm}| } 
 \hline
 \textbf{Parameters} & \textbf{Values}\\
 \hline\hline
  State & $4 \times 16 \times 16$ matrix\\
  
 Action & Up, Down, Left, Right \\
 
 Number of training episodes & $10,000$\\
 
 Episode length & $10 \times 16 \times 16$\\

 Priority replay memory size & $50,000$ \\
 
 Mini-batch size & $64$\\
 
 Discount factor & $0.90$\\
 
 Learning rate & $0.001$\\
 
 Target network update frequency & $20$\\
 
 Epsilon decay type & Exponential (Fig. \ref{fig:eps})\\
 
 Epsilon decay rate & $2100$\\
 
 Loss function & Mean Square Error\\
 
 Optimizer & Adam\\
 \hline
\end{tabular}
    \caption{List of parameters used in our experiments.}
    \label{tab:params}
\end{table}

\begin{figure}[ht!]
    \centering
    \includegraphics[width=0.8\linewidth]{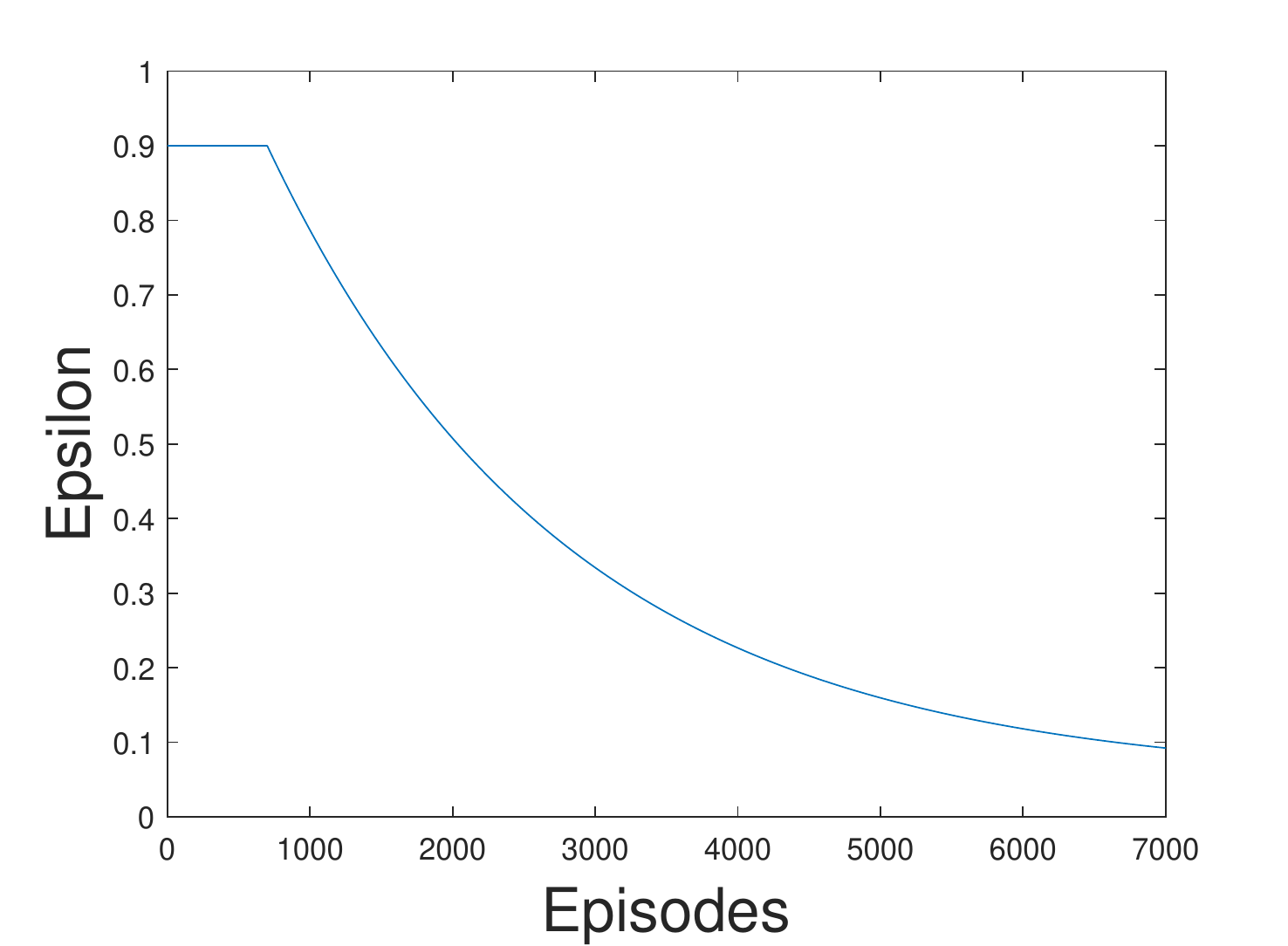}
    \caption{$\epsilon$ values used through the training episodes.}
    \label{fig:eps}
\end{figure}

\begin{figure*}[ht!]
    \centering
    \begin{tabular}{ccc}
    \includegraphics[width=0.3\linewidth]{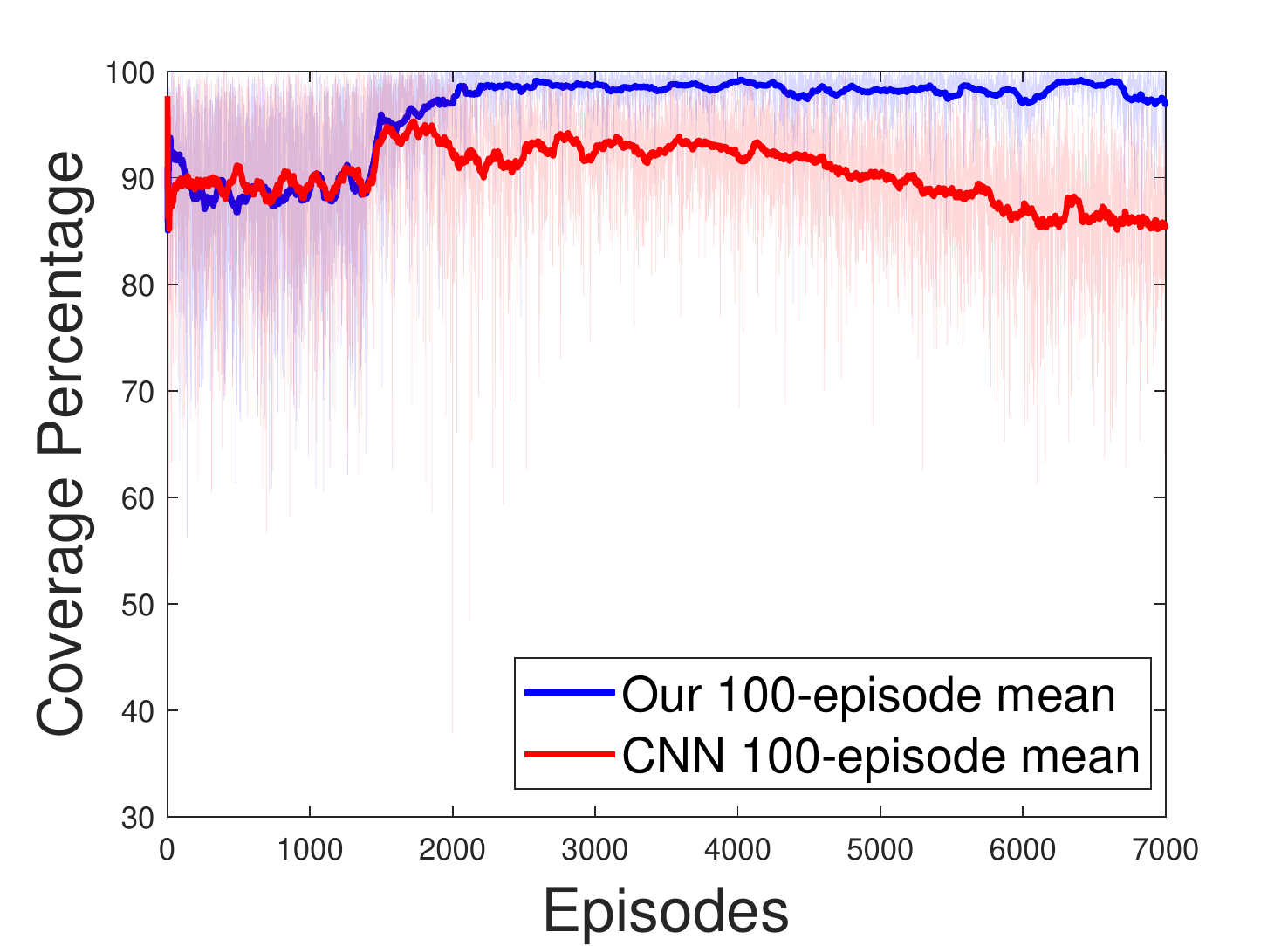} &
    \includegraphics[width=0.3\linewidth]{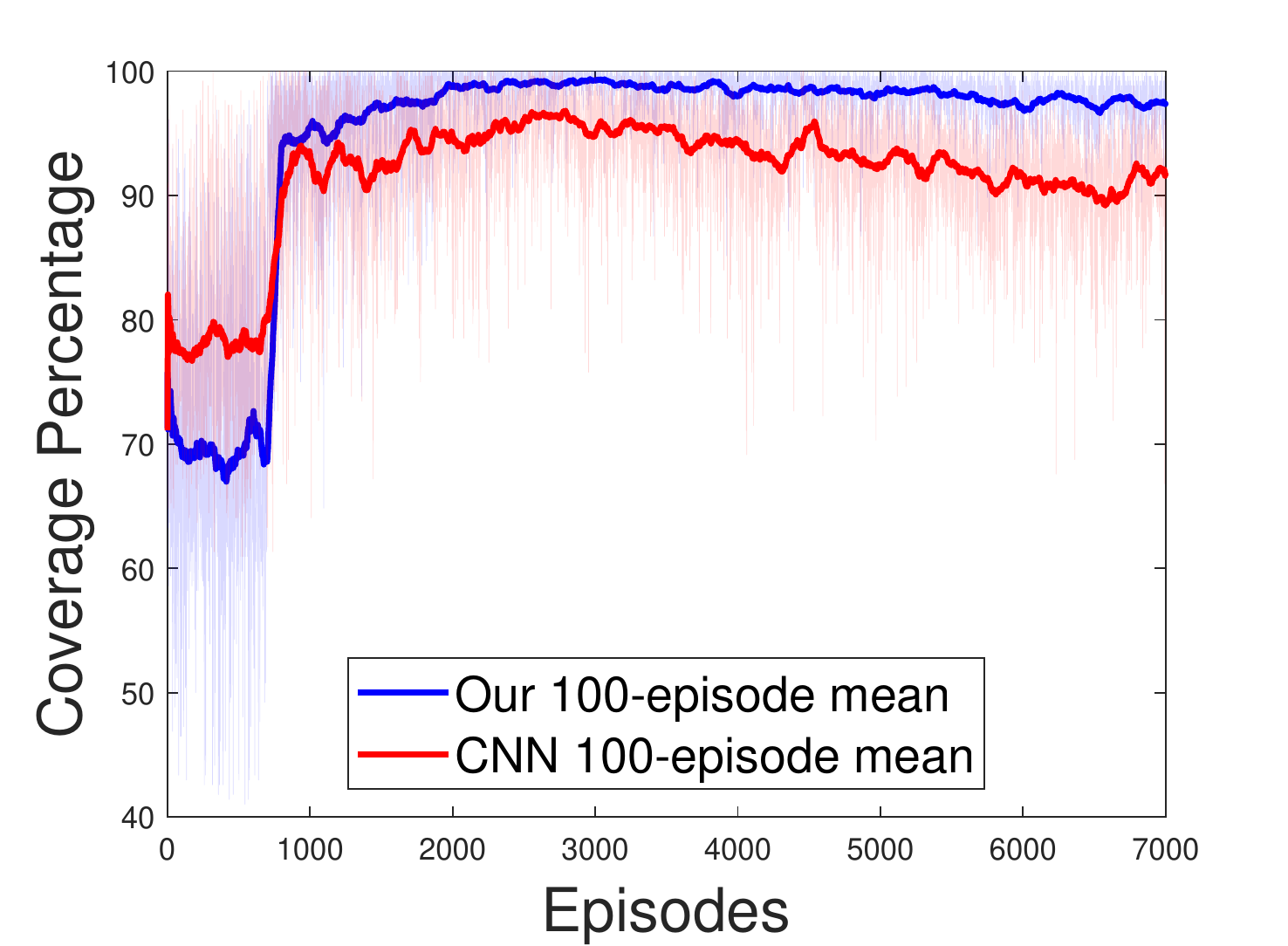}&
    \includegraphics[width=0.3\linewidth]{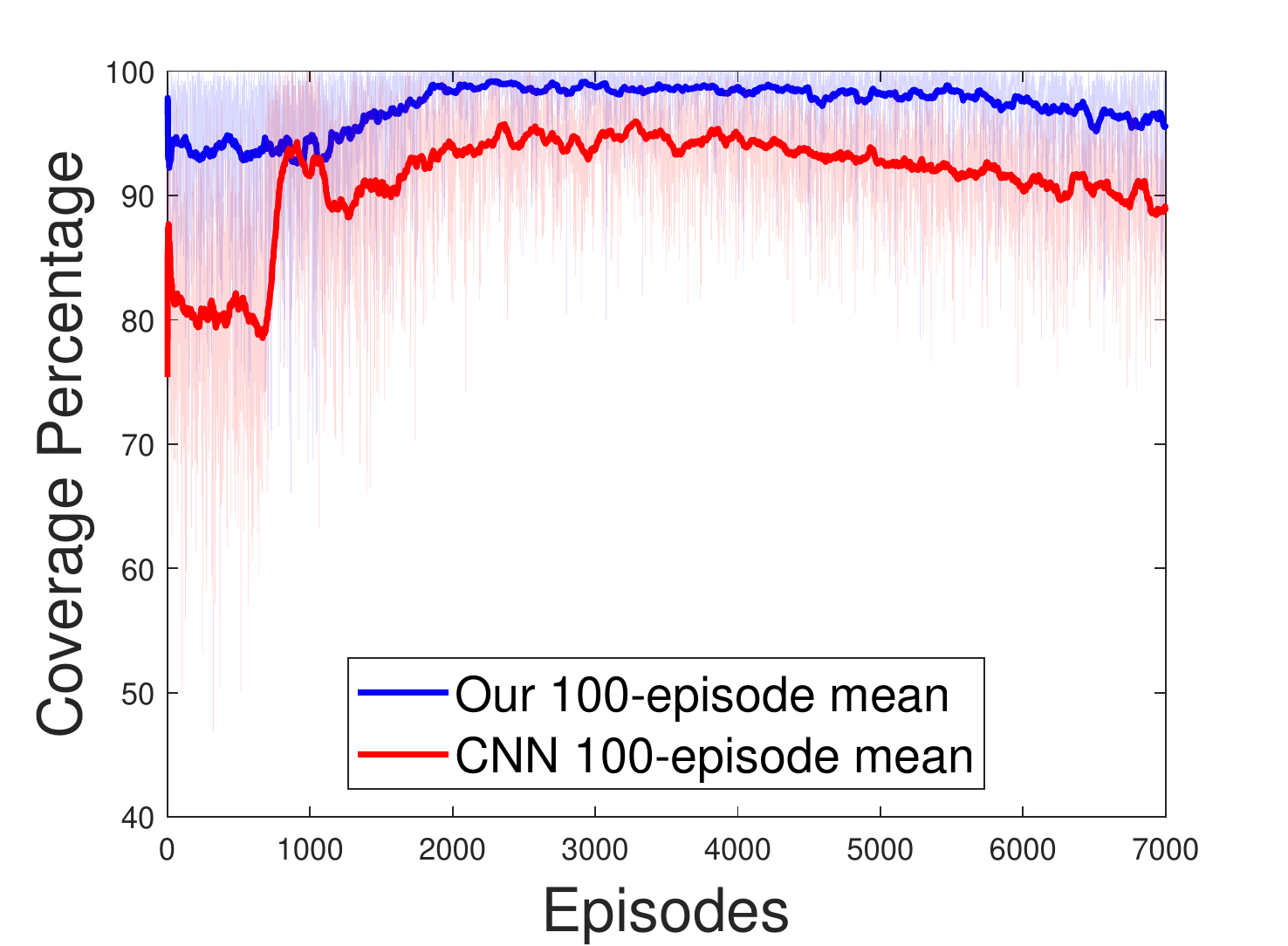}\\
    Coverage: Map 1 & Coverage: Map 2 & Coverage: Map 3\\ 
    \end{tabular}
    \caption{Percentages of coverage in three training maps.}
    \label{fig:coverage_results}
\end{figure*}

\begin{figure*}[ht!]
    \centering
    \begin{tabular}{cccc}
    \includegraphics[width=0.3\linewidth]{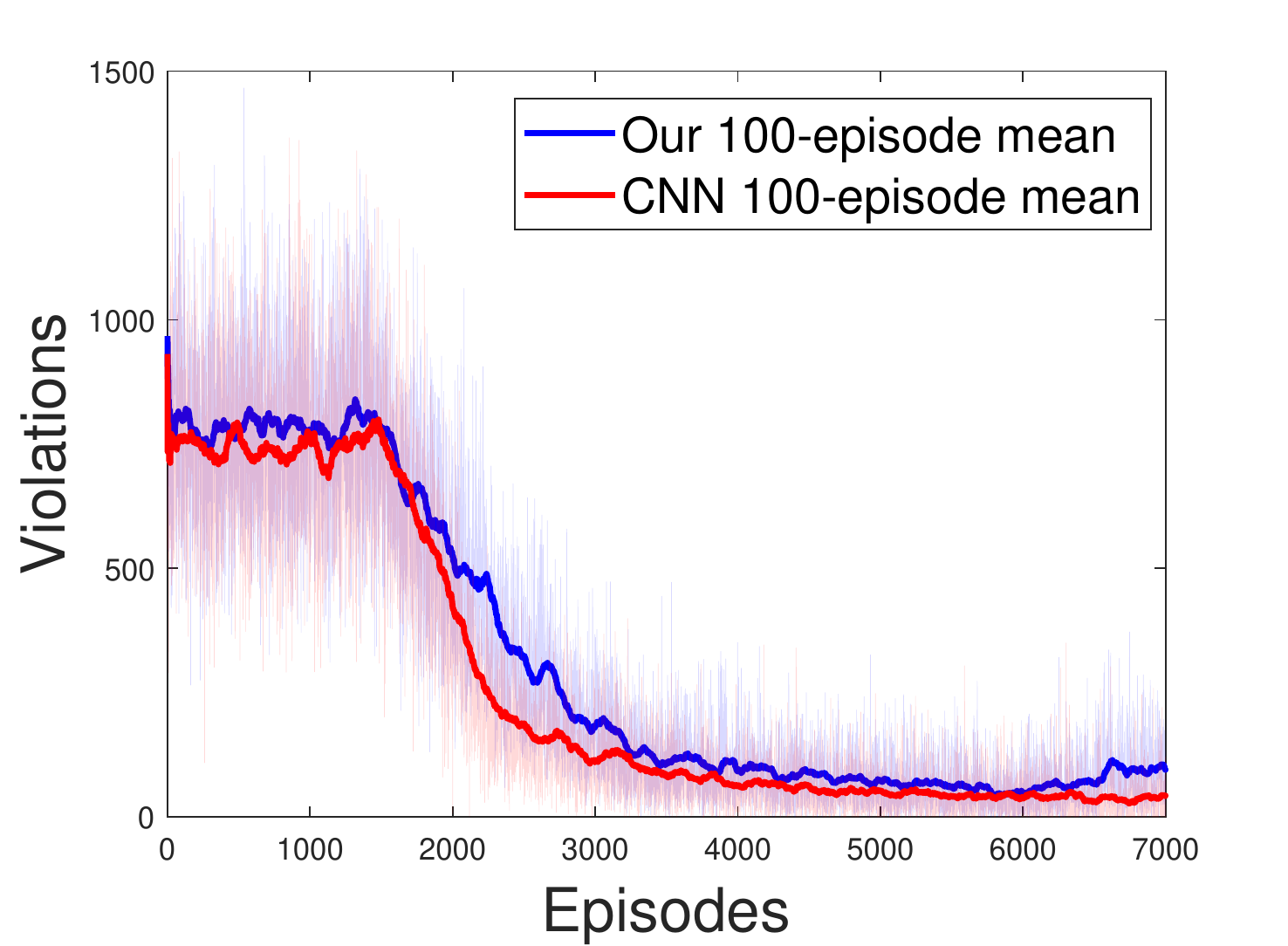} &
    \includegraphics[width=0.3\linewidth]{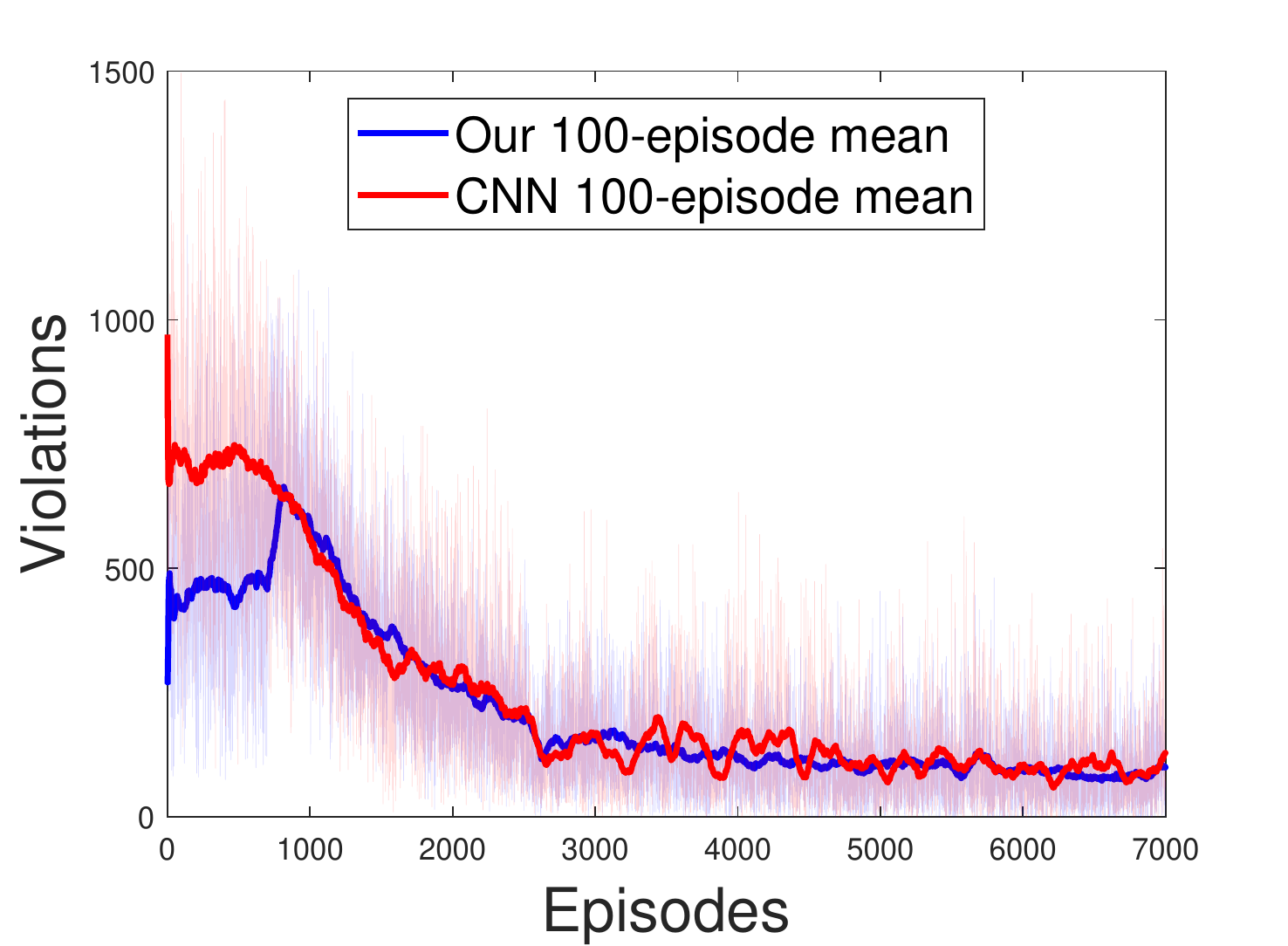}&
    \includegraphics[width=0.3\linewidth]{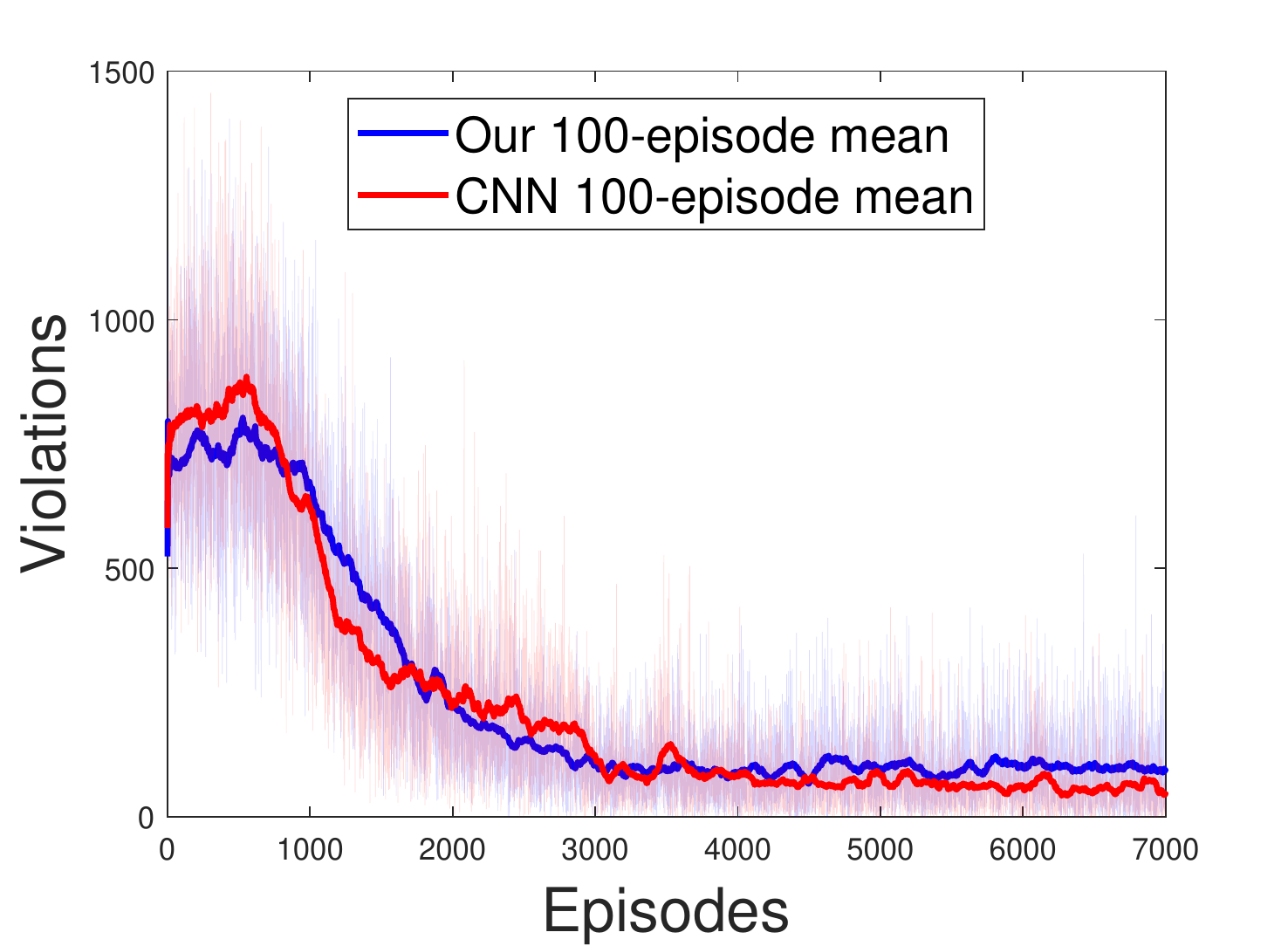}\\
    Violations: Map 1 & Violations: Map 2 & Violations: Map 3\\ 
    \end{tabular}
    \caption{The number of violations in three training maps.}
    \label{fig:viol_results}
\end{figure*}

\begin{figure*}[ht!]
    \centering
    \begin{tabular}{cccc}
    \includegraphics[width=0.3\linewidth]{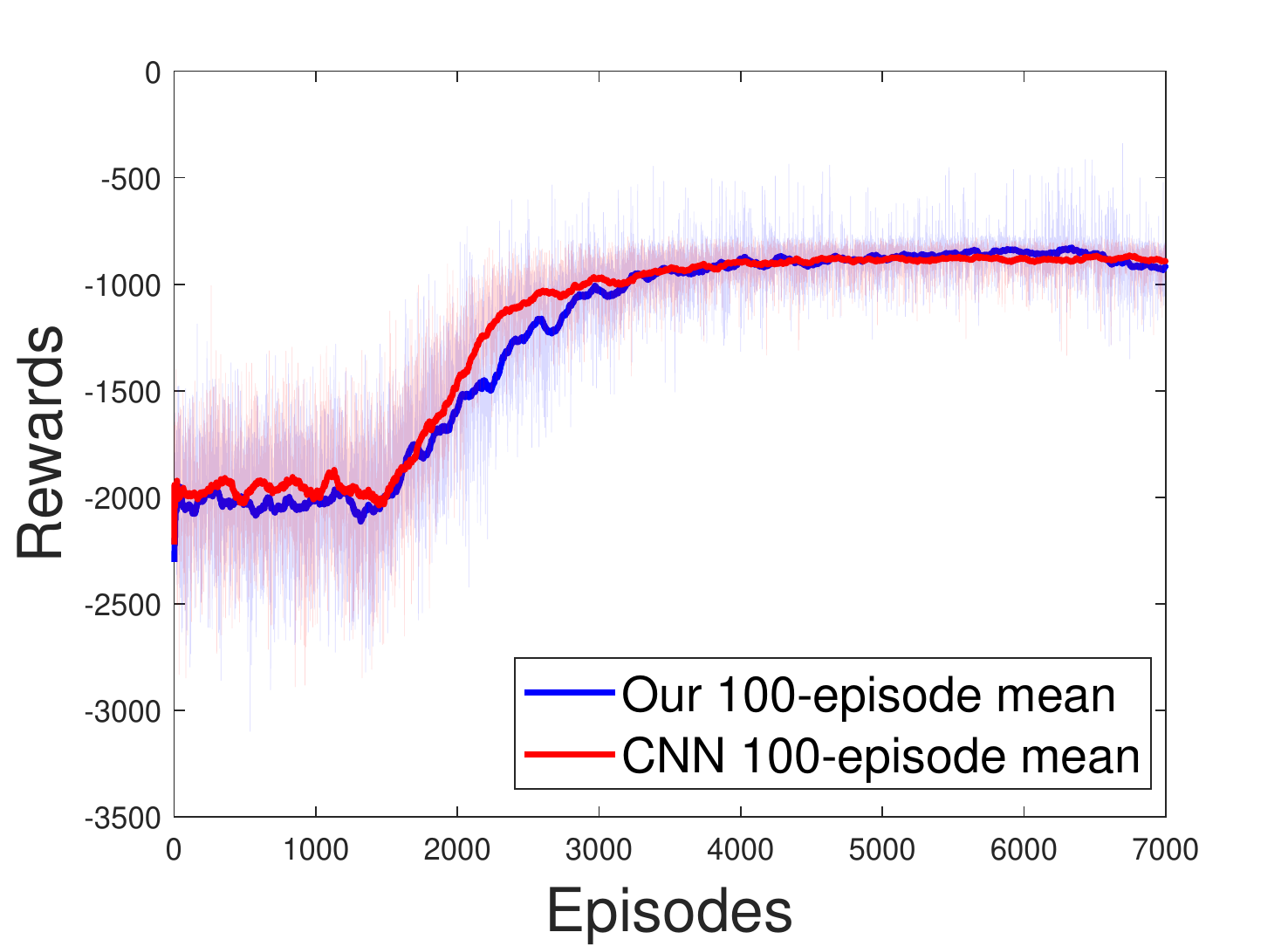} &
    \includegraphics[width=0.3\linewidth]{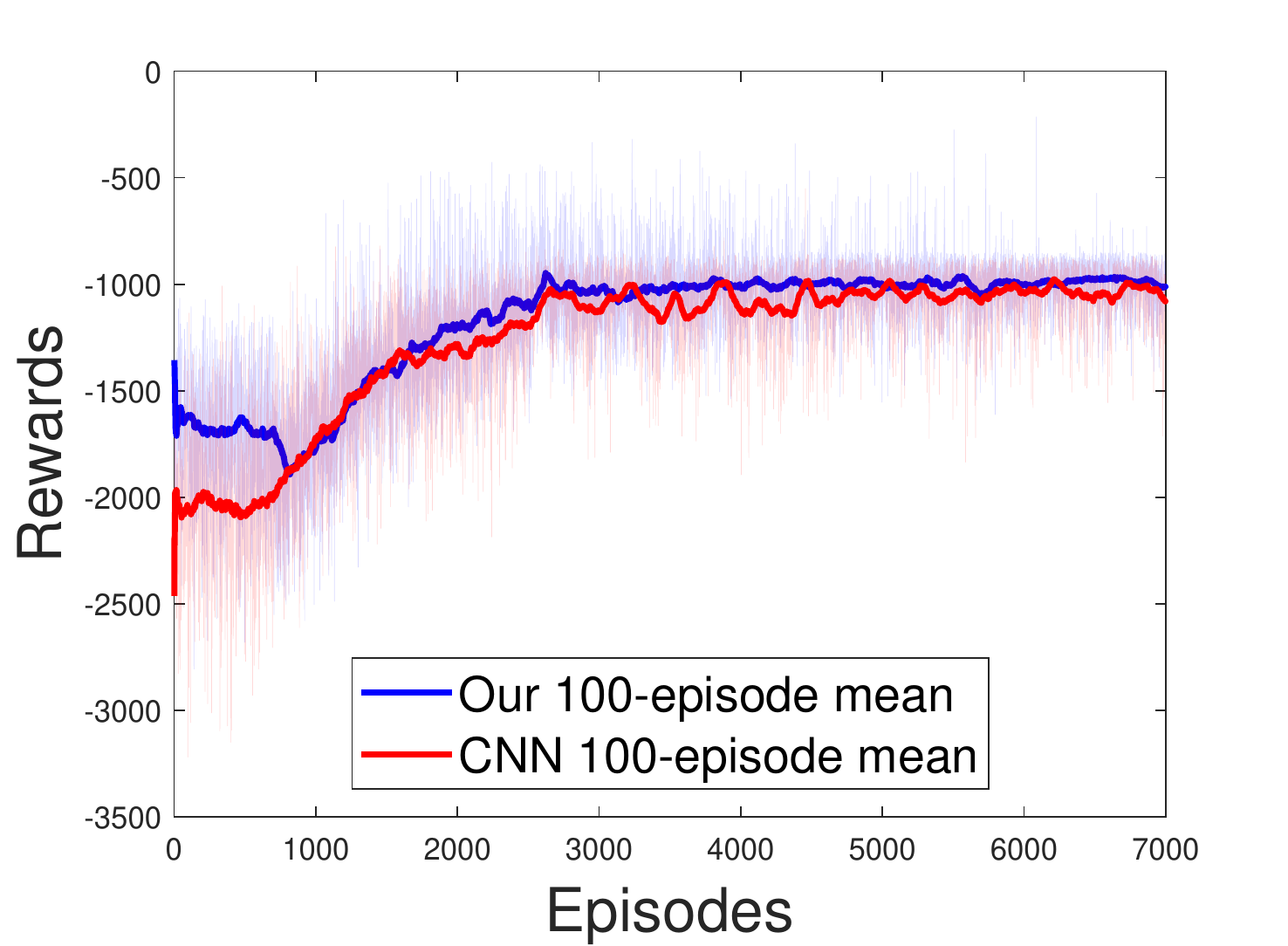} &
    \includegraphics[width=0.3\linewidth]{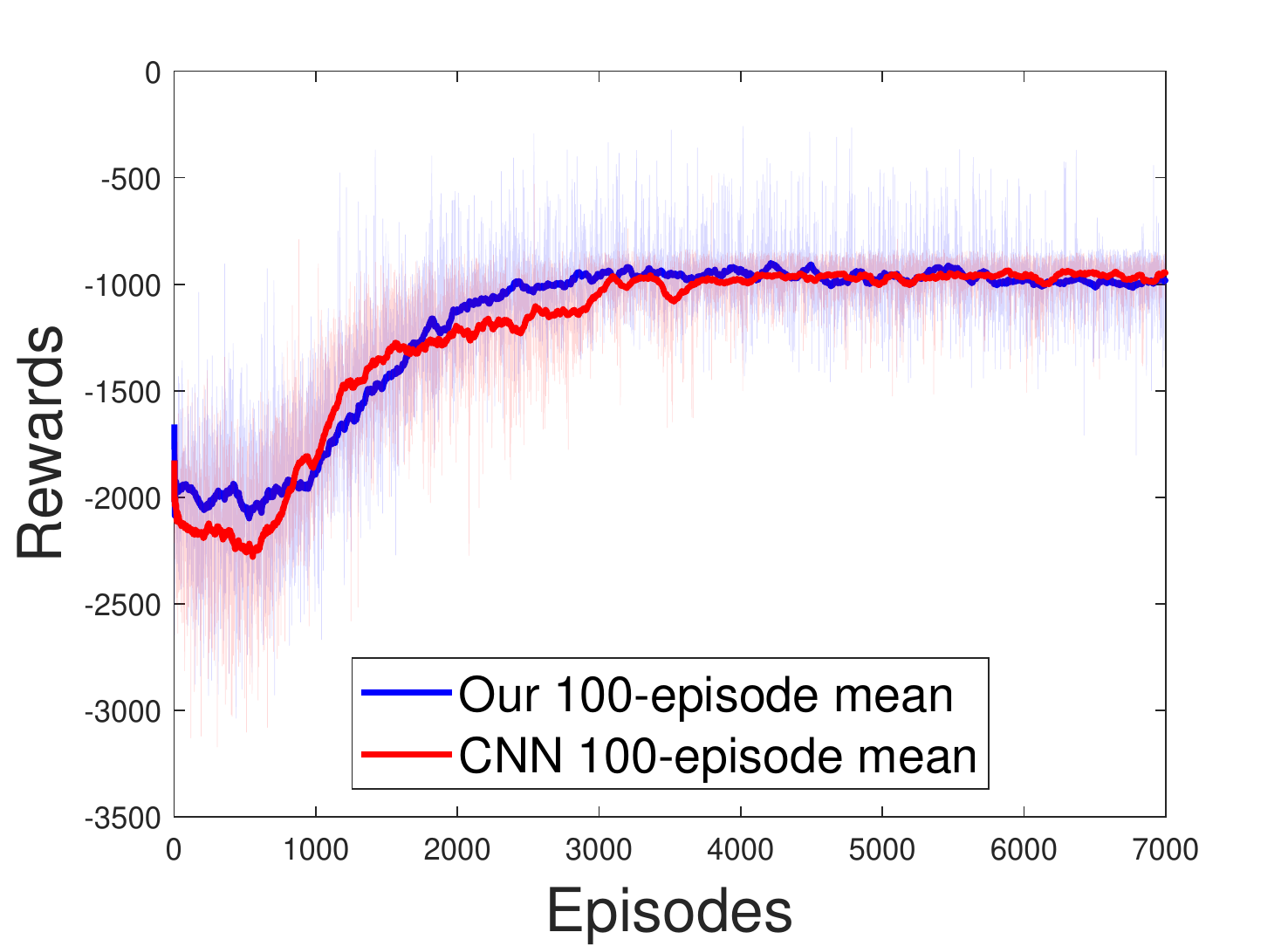}\\
    Reward: Map 1 & Reward: Map 2 & Reward: Map 3\\ 
    \end{tabular}
    \caption{Rewards in three training maps.}
    \label{fig:rwd_results}
\end{figure*}

\subsection{Results}
\begin{figure*}[ht!]
    \centering
    \begin{tabular}{ccc}
        \includegraphics[width=0.3\linewidth]{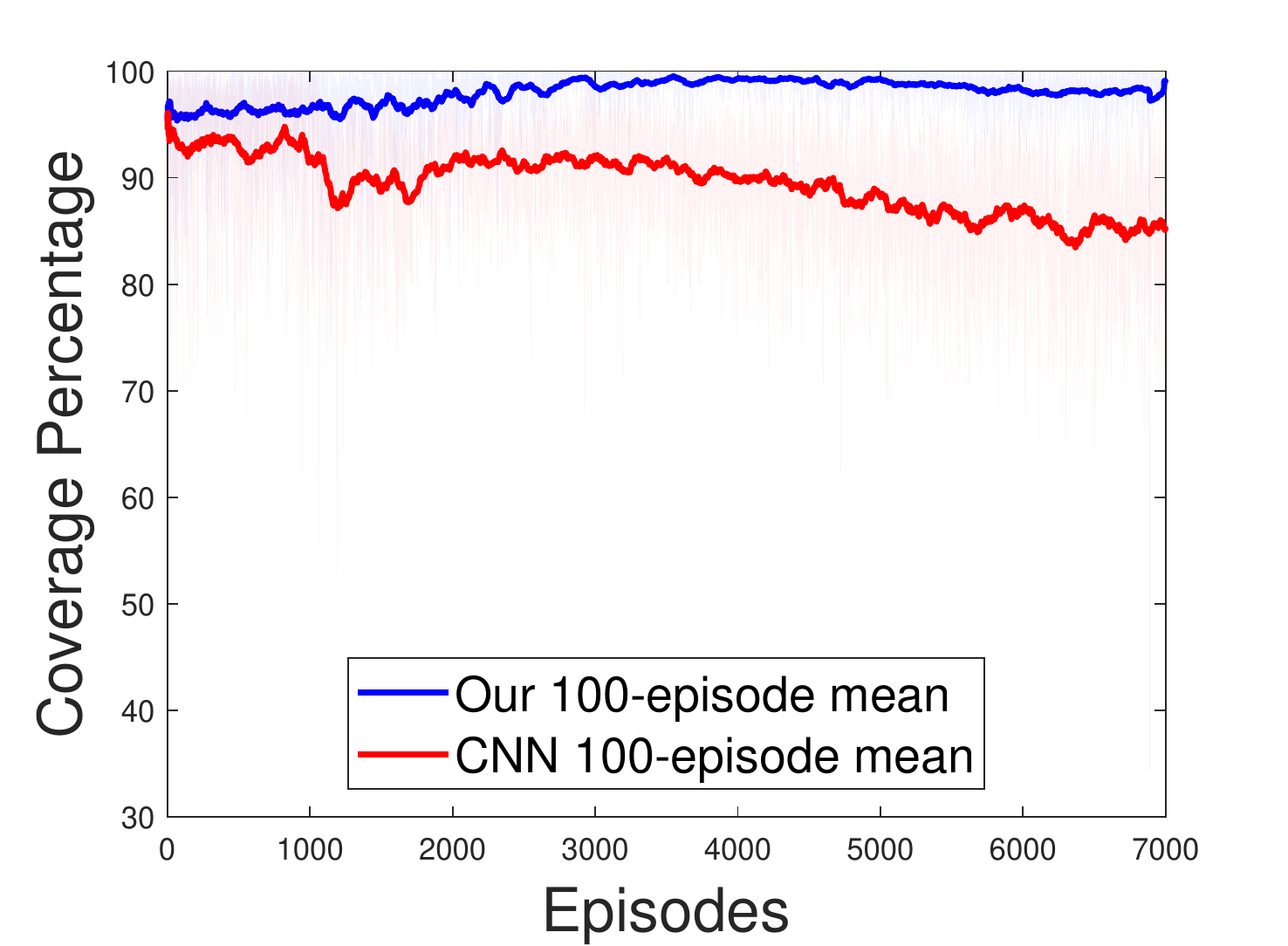} & \includegraphics[width=0.3\linewidth]{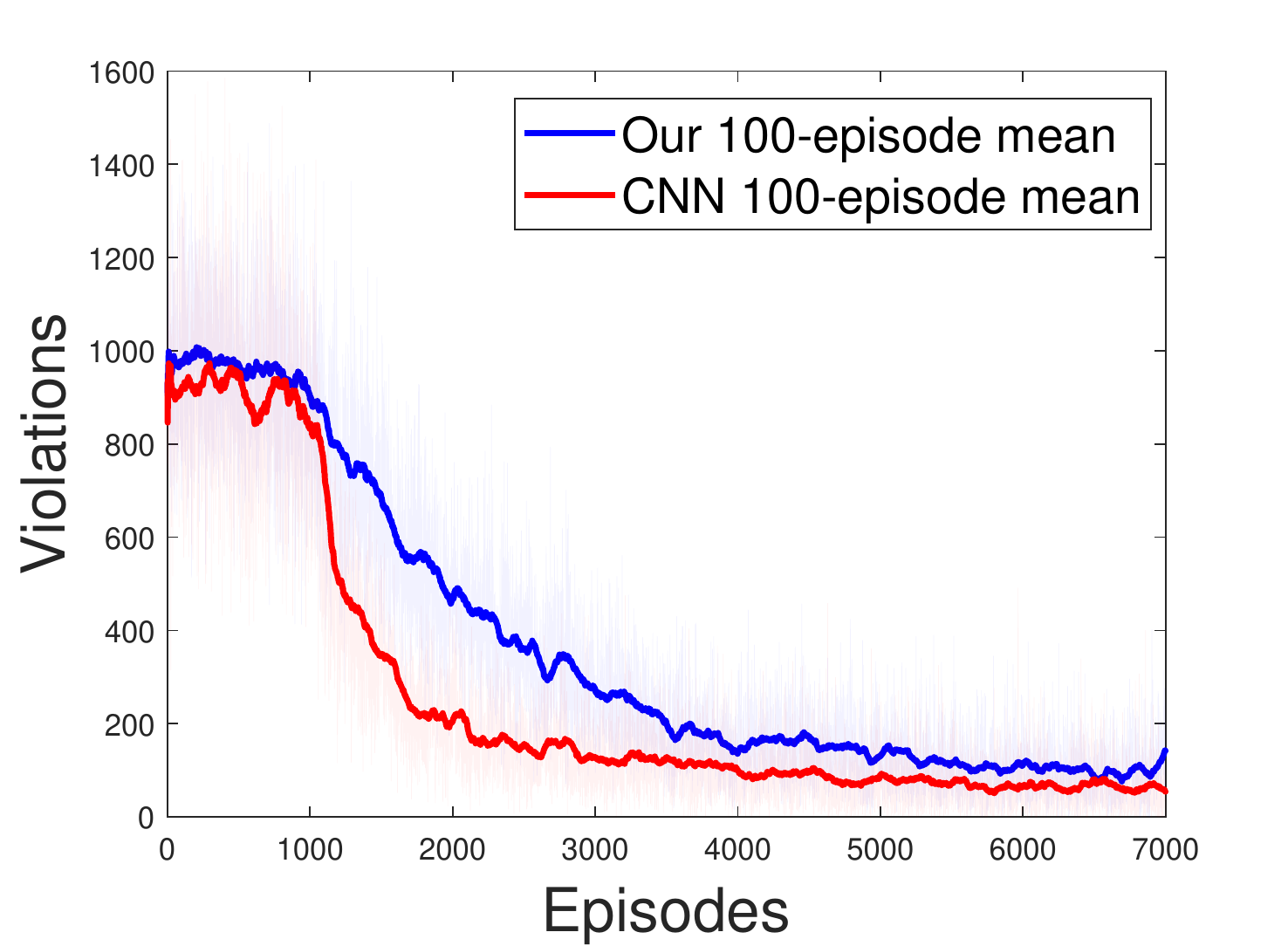} & \includegraphics[width=0.3\linewidth]{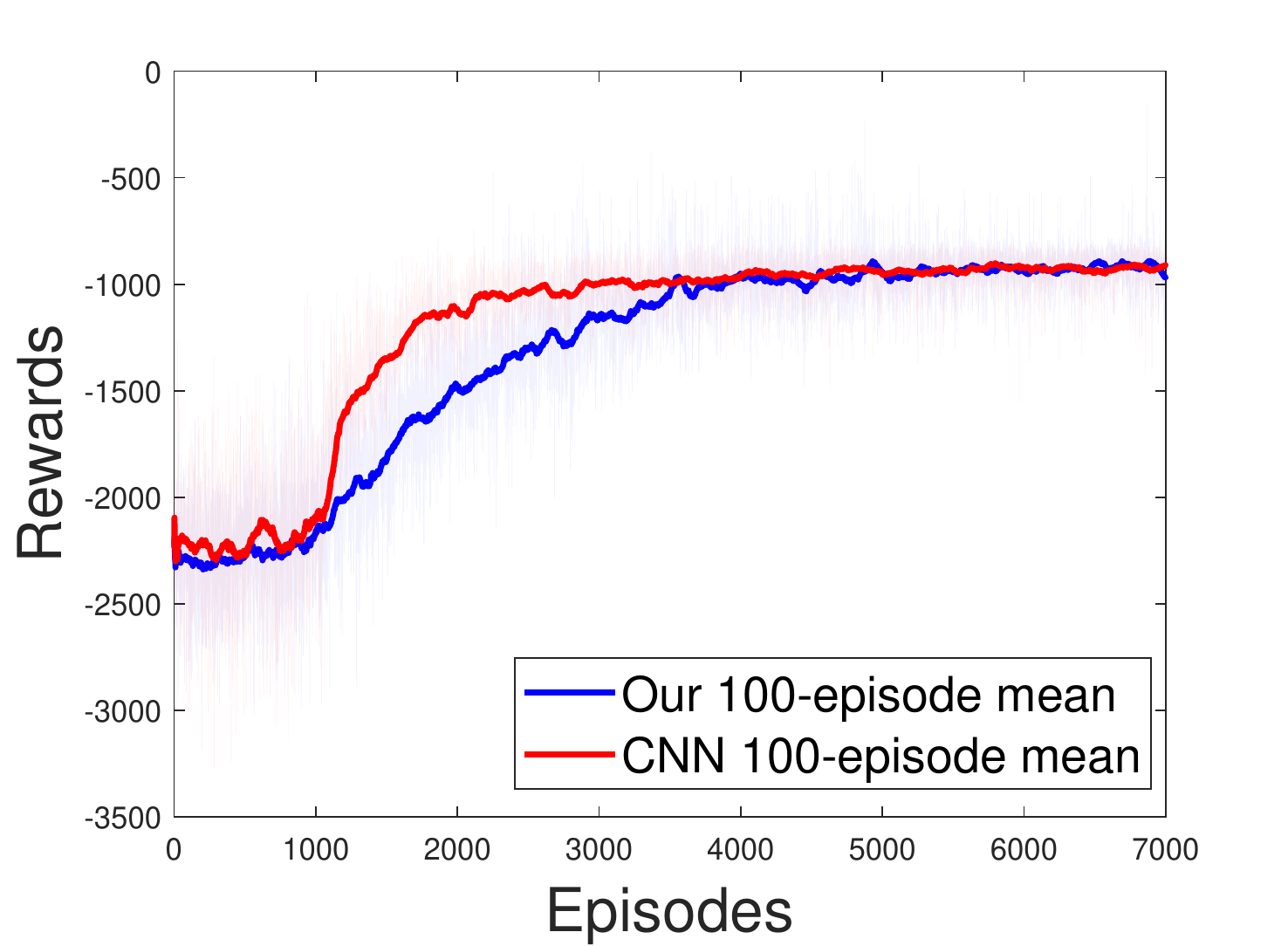}\\
    \end{tabular}
    \caption{Coverage, violation count, and reward in Map 1 with $B=4N$.}
    \label{fig:plots4n}
\end{figure*}

\begin{figure*}[ht!]
    \centering
    \begin{tabular}{ccc}
        \includegraphics[width=0.3\linewidth]{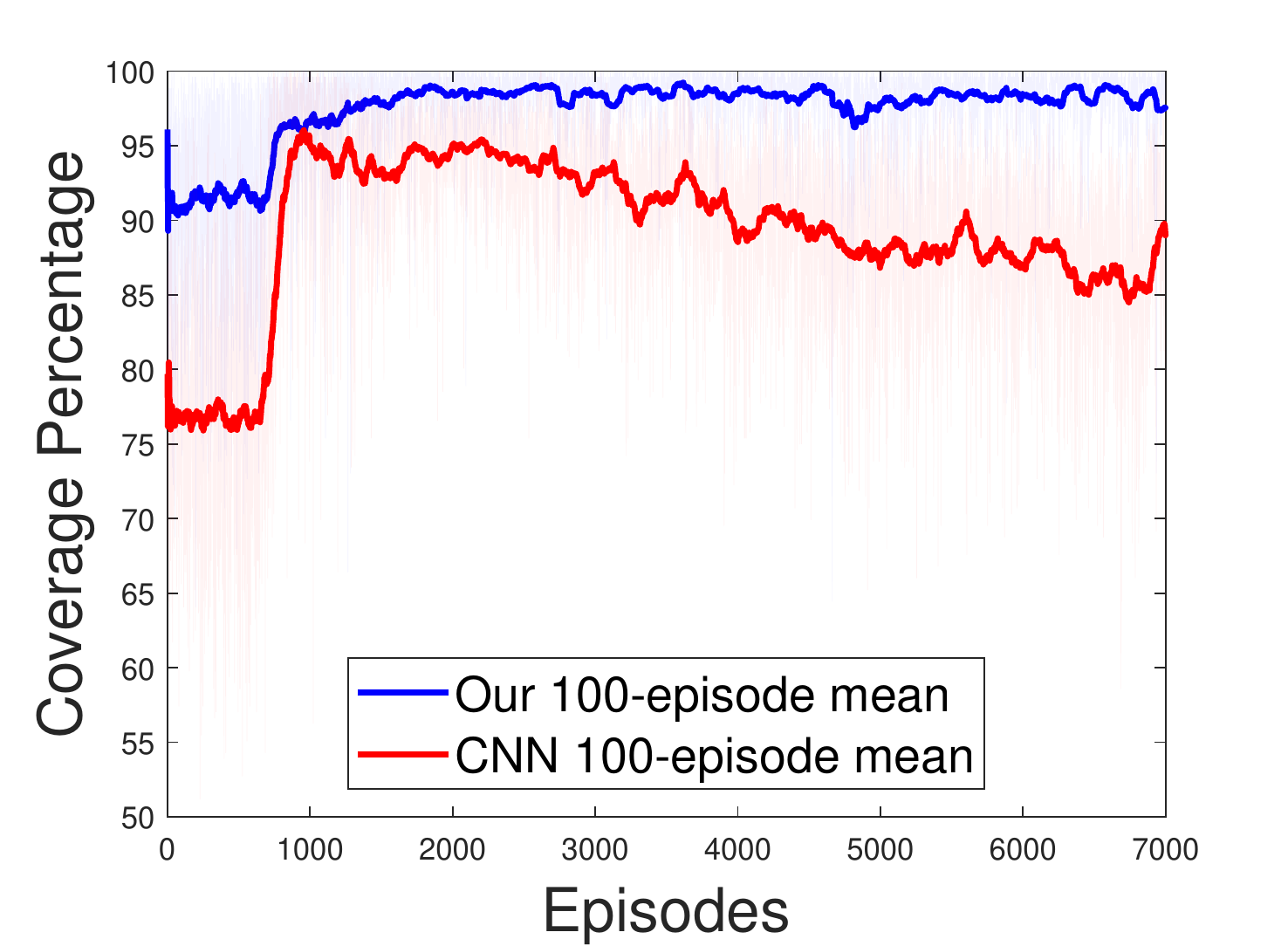} & \includegraphics[width=0.3\linewidth]{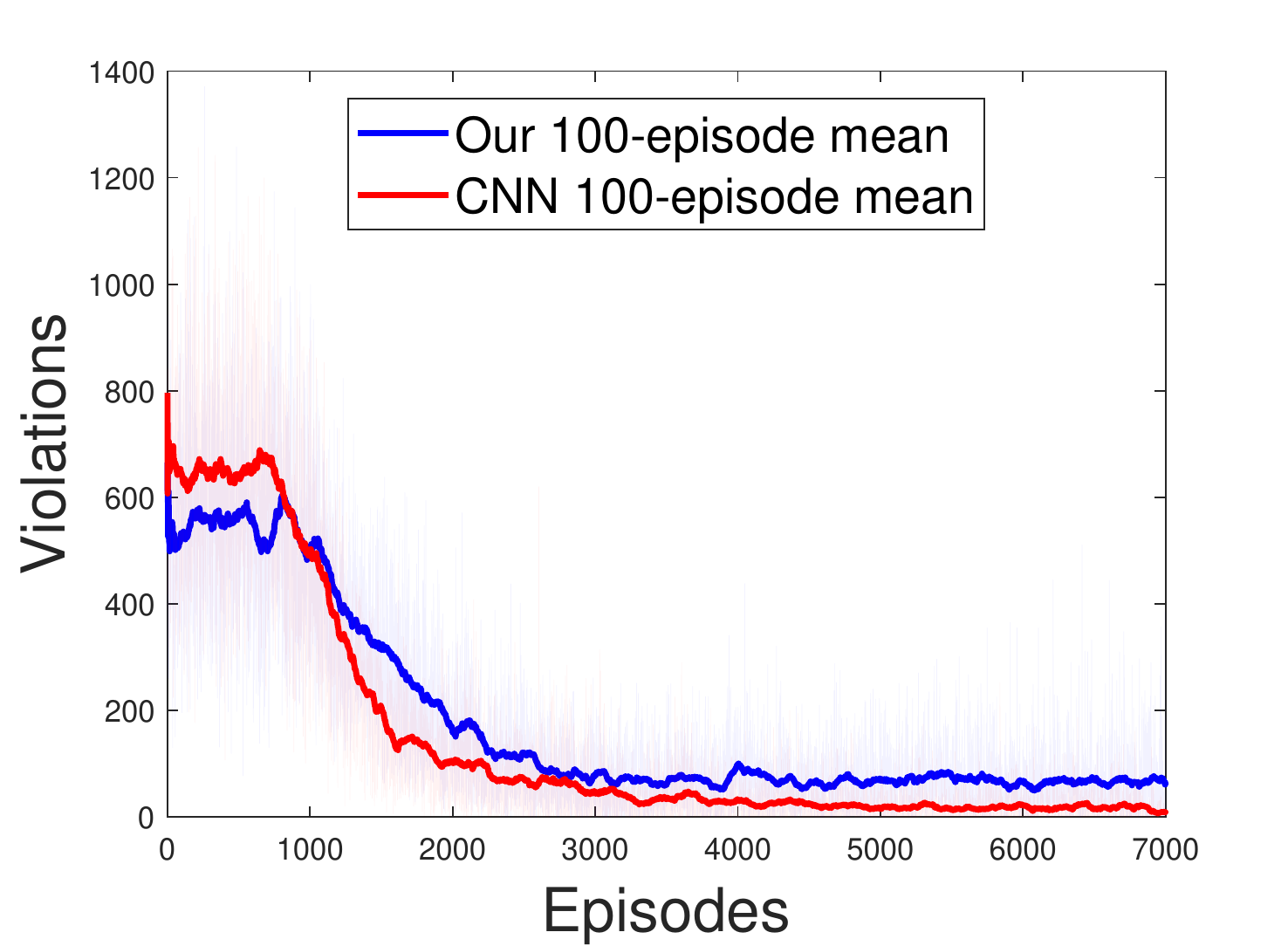} & \includegraphics[width=0.3\linewidth]{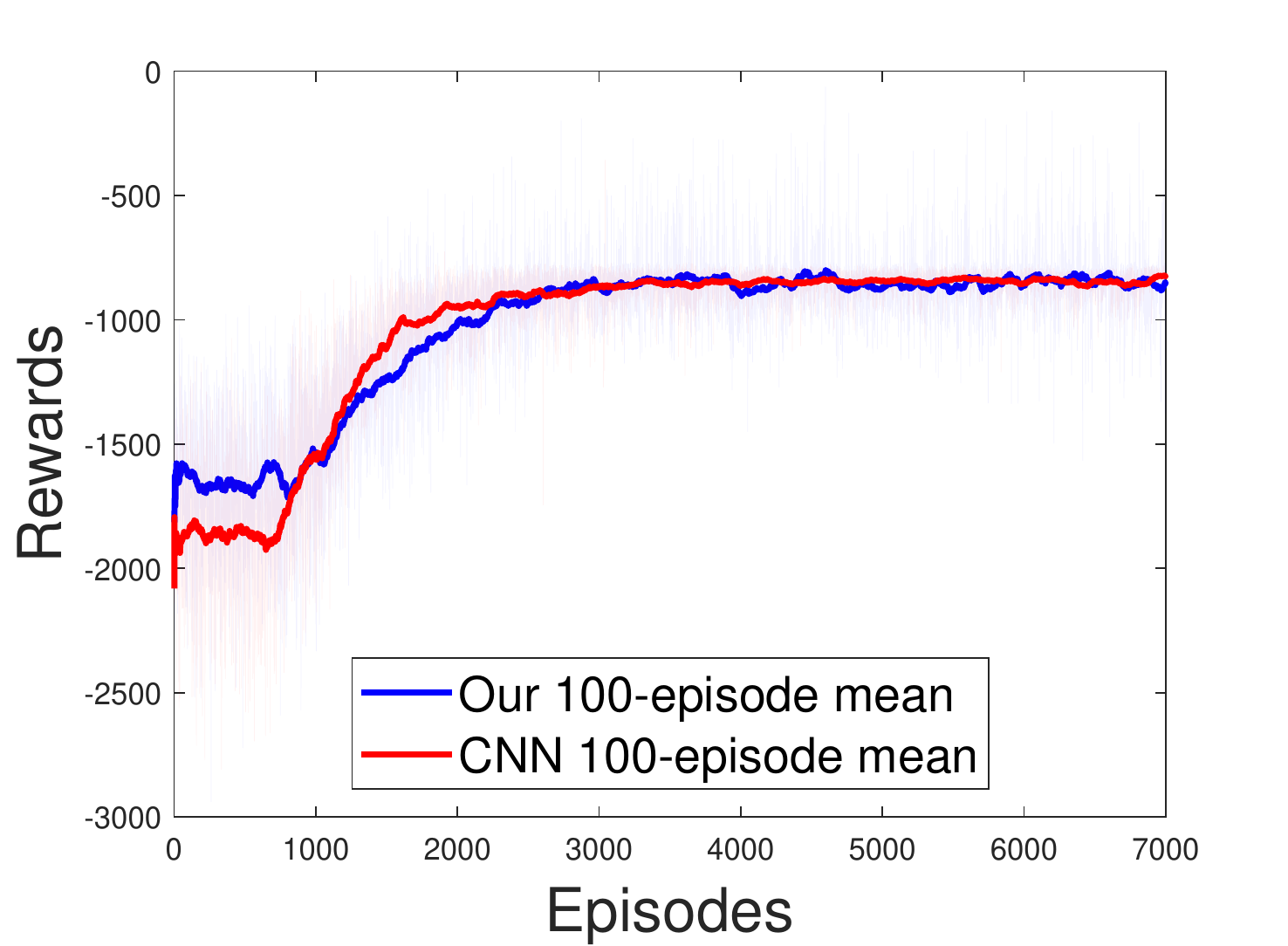}\\
    \end{tabular}
    \caption{Coverage, violation count, and reward in Map 1 with $B=6N$.}
    \label{fig:plots6n}
\end{figure*}

First, we are interested in investigating three metrics during training: coverage percentage (what percentage of cells in the environment have been covered by the robot in an episode), number of violations (the number of cells visited by the robot with negative budget in an episode), and rewards per episode. The results for these metrics for different maps are presented in Figs. \ref{fig:coverage_results}--\ref{fig:plots6n}. The faint lines in these plots indicate the raw numbers and the bold lines indicate the 100-episode moving averages.  

Results show that our proposed technique achieves higher coverage percentage than the CNN-based technique (Fig. \ref{fig:coverage_results}). Out of 7000 training episodes, our model covered the full environment 744, 1081, and 1091 times, whereas the CNN-based did this only 34, 62, and 34 times on the three tested maps respectively. On the other hand, for Maps 1 and 3, CNN yielded more episodes where $r$ did not violate the energy constraint. See Table \ref{tab:summary_result1} for reference. Overall, we are interested in finding a solution where the coverage is $100\%$ and the number of violations is 0. The episodes in which $r$ achieves both these objectives, we call them the `best' episodes. In Table \ref{tab:summary_bestRwd}, we summarize details about the best episodes yielded by the two techniques with $B=5N$. We notice that our model always yields a valid solution, whereas the CNN-based approach could never find such a solution. This highlights the advantage of using our presented model over the comparable CNN-based model. As the presence of obstacles lead the robot to take convoluted paths to cover the environment, it not only needs careful planning to cover while avoiding the obstacles, but also to recharge itself among these detours. This leads to less number of episodes with no energy constraint violation. However, due to the smaller number of free cells to cover, the robot achieves complete coverage in more episodes. These conclusions are evident from Tables \ref{tab:summary_result1}, \ref{tab:summary4N}, and \ref{tab:summary6N}, where we see that there are numerous episodes in which $r$ achieves completes coverage, but it can do so without any constraint violation in a relatively small number of episodes. Note that the `Max. Reward' row (or, column) in the tables indicate the highest reward earned by $r$ in the `best' episodes. As the episodes in which one of the objectives is not fulfilled are not candidates for the final solution, therefore, we do not report the highest rewards for them.  

\begin{table}
    \centering
    \begin{tabular}{||c|c|p{1.5cm}|p{1.5cm}||}
    \hline
        Map & Model & \# Full Cov. Episodes & \# No Vio. Episodes\\
        \hline
        \hline
        \multirow{2}{4em}{Map 1} & Our & \textbf{744} & 90 \\
        & CNN \cite{theile2020uav} & 34 & \textbf{416} \\
        \hline
        \multirow{2}{4em}{Map 2} & Our & \textbf{1081} & \textbf{108} \\
        & CNN \cite{theile2020uav} & 62 & 76 \\
        \hline
        \multirow{2}{4em}{Map 3} & Our & \textbf{1091} & 90 \\
        & CNN \cite{theile2020uav} & 34 & \textbf{191} \\
        \hline
    \end{tabular}
    \caption{Summary of performances of two individual optimization metrics.}
    \label{tab:summary_result1}
\end{table}

\begin{table}
    \centering
    \begin{tabular}{||c|c|c|c||}
    \hline
        Map & Model & \# Best Episodes & Max. Reward\\
        \hline
        \hline
        \multirow{2}{4em}{Map 1} & Our & \textbf{9} & \textbf{-339} \\
        & CNN \cite{theile2020uav} & 0 & NA \\
        \hline
        \multirow{2}{4em}{Map 2} & Our & \textbf{7} & \textbf{-215} \\
        & CNN \cite{theile2020uav} & 0 & NA \\
        \hline
        \multirow{2}{4em}{Map 3} & Our & \textbf{16} & \textbf{-259} \\
        & CNN \cite{theile2020uav} & 0 & NA \\
        \hline
    \end{tabular}
    \caption{Summary of overall performances of our approach and the baseline.}
    \label{tab:summary_bestRwd}
\end{table}

We have also tested with budget values changed to $B=4N$ and $6N$. The results are presented in Figs. \ref{fig:plots4n}, \ref{fig:plots6n} and in Tables \ref{tab:summary4N}, \ref{tab:summary6N}. Note that these experiments are run only on Map 1. When the budget is lower, the robot needs to be more innovative in planning its path that does not lead to a constraint violation. As expected from the earlier results, the CNN-based model could not find a valid solution during the training process. On the other hand, when the budget is increased to $6N$, it could find one such valid solution. However, the highest reward yielded by the final solution using our proposed technique is $5.70$ times higher than that of the CNN-based model (Table \ref{tab:summary6N}).

\begin{table}[ht!]
    \centering
    \begin{tabular}{||c|c|c||}
    \hline
        Metric & Our & CNN~\cite{theile2020uav}\\
        \hline
        \hline
        \# Full Cov. Episodes & \textbf{1097} & 28\\
        \# No Vio. Episodes & 11 & \textbf{104}\\
        \# Best Episodes & \textbf{1} & 0\\
        \# Max. Reward & \textbf{-152} & NA\\
        \hline
    \end{tabular}
    \caption{Summary of overall performances of our approach and the baseline (Map 1, $B=4N$).}
    \label{tab:summary4N}
\end{table}

\begin{table}[ht!]
    \centering
    \begin{tabular}{||c|c|c||}
    \hline
        Metric & Our & CNN~\cite{theile2020uav}\\
        \hline
        \hline
        \# Full Cov. Episodes & \textbf{1154} & 25\\
        \# No Vio. Episodes & 247 & \textbf{1568}\\
        \# Best Episodes & \textbf{55} & 1\\
        \# Max. Reward & \textbf{-63} & -358\\
        \hline
    \end{tabular}
    \caption{Summary of overall performances of our approach and the baseline (Map 1, $B=6N$).}
    \label{tab:summary6N}
\end{table}

One of the main takeaways from the results is that our proposed use of a recurrent layer leads not only to higher reward yield, but also to finding valid solutions, i.e., where the robot covers the entire environment without violating the energy constraint. Furthermore, we observe that finding such a valid solution become even more challenging with lower budget, whereas with more budget, the robot can cover the environment without budget violation relatively easily.  


\section{Conclusion and Future Work}
In this paper, we have proposed a deep reinforcement learning based solution for coverage path planning (CPP) with an energy-constrained mobile robot. Although combinatorial algorithms have been proposed for such CPP problems in the presence of a single charging station, to the best of our knowledge, these exists no solution for this problem in the presence of more than one charging station. The environment is discretized into square grid cells and the robot is allowed to move to any of the four orthogonal neighbor cells at any given step. Our proposed solution in this paper uses a deep Q-network based technique that employs two convolutional neural layer along with a recurrent neural module, called Long Short Term Memory (LSTM). The proposed neural network architecture takes state inputs via four channels and outputs Q-values for the possible actions. We have compared our proposed technique against a comparable solution that does not use any recurrent neural component. We trained the model on three environments with different obstacle configurations and with various budgets. Results show that our proposed technique outperforms the baseline in all the used environment configurations. 

Given that this is the first study on CPP in the presence of multiple charging stations, our assumption was that the environment configuration, i.e., the locations of the obstacles are known \textit{a priori}, and therefore, the spatial map of these obstacles can be passed as part of the state input to the neural network. In the future, we plan to extend this work to incorporate unknown obstacles. One potential direction would be to use the robot's on-board sensors such as a laser range finder to detect the nearby obstacles and build the obstacle map on-the-fly. This partial information can later be fed to the neural network. Using a recurrent module such as an LSTM, the robot will learn to avoid the obstacles while covering the entire environment without violating the energy constraint. Another future direction of investigation would be to consider dynamic obstacles unlike the static ones that have usually been considered in related studies.

\section*{Declarations}
The datasets generated during and/or analysed during the current study are available from the corresponding author on reasonable request.

\bibliography{references}

\end{document}

%% file: init.tex
\usetikzlibrary{quotes,arrows.meta}
\usetikzlibrary{positioning}

\def\edgecolor{rgb:blue,4;red,1;green,4;black,3}
\newcommand{\midarrow}{\tikz \draw[-Stealth,line width =0.8mm,draw=\edgecolor] (-0.3,0) -- ++(0.3,0);}

\usepackage{Ball}
\usepackage{Box}
\usepackage{RightBandedBox}

%% file: statespace_illust.tex

\usetikzlibrary{positioning}

\begin{tikzpicture}[scale=.45,every node/.style={minimum size=1cm},on grid]
		
    \begin{scope}[
    	yshift=-166,every node/.append style={
    	yslant=0.5,xslant=-1},yslant=0.5,xslant=-1]
    	\fill[white,fill opacity=0.9] (0,0) rectangle (8,8);
        \draw[gray,very thin] (0,0) grid (8,8); 
        \draw[black,very thick] (0,0) rectangle (8,8);
        \fill[orange,fill opacity=.5] (0.05,0.05) rectangle (1,1);
        \node at (0.5,0.5) {$1$};
        \fill[orange,fill opacity=.5] (0,2) rectangle (1,1);
        \node at (0.5,1.5) {$1$};
        \fill[orange,fill opacity=.5] (2,0) rectangle (1,1);
        \node at (1.5,0.5) {$1$};
    \end{scope}
    
    \begin{scope}[
            yshift=-83,every node/.append style={
            yslant=0.5,xslant=-1},yslant=0.5,xslant=-1
            ]
        \fill[white,fill opacity=0.9] (0,0) rectangle (8,8);
        \draw[gray,very thin] (0,0) grid (8,8); 
        \draw[black,very thick] (0,0) rectangle (8,8);
        \fill[red,fill opacity=.5] (2,2) rectangle (1,1);
        \node at (1.5,1.5) {$1$};
    \end{scope}
    	
    \begin{scope}[
    	yshift=0,every node/.append style={
    	    yslant=0.5,xslant=-1},yslant=0.5,xslant=-1
    	             ]
        \fill[white,fill opacity=.9] (0,0) rectangle (8,8);
        \draw[black,very thick] (0,0) rectangle (8,8);
        \draw[gray,very thin] (0,0) grid (8,8);
        \fill[green,fill opacity=.5] (0.05,0.05) rectangle (1,1);
        \fill[green,fill opacity=.5] (6,3) rectangle (7,4);
        \fill[green,fill opacity=.5] (3,7) rectangle (4,8);
        \node at (0.5,0.5) {$1$};
        \node at (6.5,3.5) {$1$};
        \node at (3.5,7.5) {$1$};
    \end{scope}
    
       \begin{scope}[
    	yshift=83,every node/.append style={
    	    yslant=0.5,xslant=-1},yslant=0.5,xslant=-1
    	             ]
        \fill[white,fill opacity=.6] (0,0) rectangle (8,8);
        \draw[black,very thick] (0,0) rectangle (8,8);
        \draw[black] (0,0) grid (8,8);
        \fill[fill=black!40!white, fill opacity=.5] (2,2) rectangle (4,4);
        \fill[fill=black!40!white, fill opacity=.5] (4,7) rectangle (8,8);
        \node at (2.5,2.5) {$1$};
        \node at (3.5,2.5) {$1$};
        \node at (3.5,3.5) {$1$};
        \node at (2.5,3.5) {$1$};
        
        \node at (4.5,7.5) {$1$};
        \node at (5.5,7.5) {$1$};
        \node at (6.5,7.5) {$1$};
        \node at (7.5,7.5) {$1$};
    \end{scope}

    	



    \draw[-latex,thick,gray](6.6,5)node[right]{Obstacles}
        to[out=180,in=90] (4.1,5);
        
    \draw[-latex,thick,green] (4.2,2) node[right]{Charging stations}
         to[out=180,in=90] (2,2);

    \draw[-latex,thick,red](4,-.3)node[right]{Current location}
        to[out=180,in=90] (1.9,-1);
        
        \draw[-latex,thick,orange](3.9,-3)node[right]{Past locations}
        to[out=180,in=90] (2,-3.5);





\end{tikzpicture}

%% file: NN_model.tex
\def\ConvColor{rgb:yellow,5;red,1.5;white,5}
\def\ConvReluColor{rgb:yellow,5;red,15;white,5}
\def\PoolColor{rgb:red,1;black,0.3}
\def\DcnvColor{rgb:blue,5;green,2.5;white,5}
\def\SoftmaxColor{rgb:magenta,5;black,7}
\def\SumColor{rgb:blue,5;green,15}
\def\poolsep{1}

\tikzset{
connector/.style={
     -latex,
     font=\scriptsize
    },
rectangle connector/.style={
        connector,
        to path={(\tikztostart) -- ++(#1,0cm) \tikztonodes |- (\tikztotarget) },
        pos=0.5
    },
    rectangle connector/.default=-2cm,
    straight connector/.style={
        connector,
        to path=--(\tikztotarget) \tikztonodes
    }
    }
\tikzset{every loop/.style={min distance=20mm
}}

\begin{tikzpicture}
\tikzstyle{connection}=[thick,every node/.style={sloped,allow upside down},draw=\edgecolor,opacity=0.6]
\pic[shift={(0,0,0)}] at (0,0,0) {RightBandedBox={name=cr1,caption=Conv 1,%
        fill=\ConvColor,bandfill=\ConvReluColor,%
        height=10,width={1},depth=10}};
\pic[shift={(1,0,0)}] at (cr1-east) {RightBandedBox={name=cr2,caption=Conv 2,%
        fill=\ConvColor,bandfill=\ConvReluColor,%
        height=10,width={1},depth=10}};


\pic[shift={(-2,0,0)}] at (cr1-west) {Box={name=states,caption=State Input,%
xlabel={{"","dummy"}},fill=,opacity=0.01,height=5,width=3,depth=5}};




\pic[shift={(2.5,0,0)}] at (cr2-east) {Box={name=lstm,caption=LSTM,%
        fill=,opacity=0.5,height=5,width={7},depth=3,fill=\SoftmaxColor}};
        
\node[circle, draw=red] (anch) [left=of lstm-west] {$+$};

\node[draw=none] (budget) [below=of anch,yshift=-0.2cm] {Budget};
        
\def\skipshift{6.5}
\pic[shift={(1.5,0,0)}] at (lstm-east) {Box={name=lin2,caption=Linear, 
zlabel={$\qquad 128$},
fill=\SumColor,opacity=0.3,height=2,width=2,depth=15}};

\pic[shift={(1.5,0,0)}] at (lin2-east) {Box={name=output,caption=Output,%
        xlabel={{"$\vert\mathcal{A}\vert$","dummy"}},fill=\DcnvColor,height=10,width=1,depth=0,zlabel=}};
\draw [connection]  (states-east)      -- node {\midarrow} (cr1-west);

\draw [connection]  (cr1-east)      -- node {\midarrow} (cr2-west);
\draw [connection]  (cr2-east)    -- node {\midarrow} (anch);
\draw [connection]  (budget)    -- node {\midarrow} (anch);
\draw [connection]  (anch)    -- node {\midarrow} (lstm-west);
\draw [connection]  (lstm-east)    -- node {\midarrow} (lin2-west);
\draw [connection]  (lin2-east)    -- node {\midarrow} (output-west);

\end{tikzpicture}